\documentclass[journal]{IEEEtran}
\usepackage{bm}
\usepackage{multirow}
\usepackage{graphicx}
\usepackage{color}
\usepackage{amsmath,amssymb}
\usepackage[hidelinks]{hyperref}
\usepackage{caption}
\usepackage{subcaption}
\ifCLASSINFOpdf
\else
\fi

\newcommand{\riii}[1]{\textcolor{black}{#1}}
\newcommand{\new}[1]{#1}

\begin{document}
\title{Video Crowd Localization with Multi-focus Gaussian Neighborhood Attention and a Large-Scale Benchmark}

\author{Haopeng~Li,
        Lingbo~Liu,
        Kunlin~Yang,
        Shinan~Liu,
        Junyu~Gao,
        Bin~Zhao,
        Rui~Zhang 
        and~Jun~Hou
\thanks{Haopeng Li is with the School of Computing and Information Systems, University of Melbourne. E-mail: haopeng.li@student.unimelb.edu.au.}% <-this % stops a space
\thanks{Lingbo Liu is with the School of Computer Science and Engineering, Sun Yat-sen University. E-mail: liulingbo918@gmail.com.}
\thanks{Kunlin Yang, Shinan Liu, and Jun Hou are with SenseTime Group Limited. E-mail: \{yangkunlin,liushinan,houjun\}@sensetime.com.}
\thanks{JunyuGao and Bin Zhao are with the School of Artificial Intelligence, Optics and Electronics, Northwestern Polytechnical University. E-mail: \{gjy3035,binzhao111\}@gmail.com.}
\thanks{Rui Zhang (\url{www.ruizhang.info}) is with Tsinghua University. E-mail: rayteam@yeah.net.}
%\thanks{Corresponding author: Kunlin Yang.}
%\thanks{$^*$Work done during an internship at SenseTime Group Limited.}
}

\maketitle

\begin{abstract}
Video crowd localization is a crucial yet challenging task, which aims to estimate exact locations of human heads in the given crowded videos. To model spatial-temporal dependencies of human mobility, we propose a multi-focus Gaussian neighborhood attention (GNA), which can effectively exploit long-range correspondences while maintaining the spatial topological structure of the input videos. In particular, our GNA can also capture the scale variation of human heads well using the equipped multi-focus mechanism.  Based on the multi-focus GNA, we develop a unified neural network called GNANet to accurately locate head centers in video clips by fully aggregating spatial-temporal information via a scene modeling module and a context cross-attention module. Moreover, to facilitate future researches in this field, we introduce a large-scale crowd video benchmark named VSCrowd\footnote{\url{https://github.com/HopLee6/VSCrowd}}, which consists of 60K+ frames captured in various surveillance scenes and 2M+ head annotations. Finally, we conduct extensive experiments on three datasets including our VSCrowd, and the experiment  results show that the proposed method is capable to achieve state-of-the-art performance for both video crowd localization and counting. 

\end{abstract}
% \footnote{The code and annotated video examples can be found in supplemental materials.}


% \fi
\IEEEpeerreviewmaketitle

\section{Introduction}
% \IEEEPARstart{D}{ue} to its important applications such as public safety, surveillance-based crowd analysis \cite{zhan2008crowd,sohn2020laying,kang2018beyond} has drawn widespread attention in both academic and industrial communities. 

\IEEEPARstart{S}{urveillance}-based crowd analysis \cite{zhan2008crowd,sohn2020laying,kang2018beyond} has drawn widespread attention in both academic and industrial communities due to its important applications in public safety.
Over the past decade, numerous models \cite{wang2020nwpu,liu2019crowd,cao2018scale,liu2019context,zhang2019attentional,yan2019perspective,luo2020hybrid,liu2020weighing,sindagi2019ha} were proposed for crowd counting, but most of them only generated coarse density maps whose summations are the counts. Although achieving great performance in crowd counting, they cannot obtain accurate locations of human heads, which limits their applications for many more delicate and valuable downstream tasks such as crowd tracking and crowd prediction. For example, crowd tracking requires the exact locations of all people at arbitrary history time to recover the trajectories of people in a video sequence. These tasks are of great importance to public safety. However, most of the crowd counting methods generate coarse density maps reflecting the spatial distribution of the crowd, and such density maps are insufficient for those downstream tasks. Motivated by this, we focus on a more fundamental but challenging task, crowd localization, which targets to automatically locate the human heads of various scales in unconstrained scenes.

Intuitively, crowd localization can be modeled as head detection and addressed using common object detection frameworks such as the one-stage detectors (FCOS \cite{tian2019fcos}, YOLO \cite{redmon2016you}, {etc}.) and  the two-stage detectors  (Faster-RCNN \cite{ren2015faster}, RFCN \cite{dai2016r}, {etc}.). However, these detection methods are not capable of handling extremely crowd scenes where the human heads are often tiny and overlapped with each other \cite{zou2019object}. As for the face-specific detection methods \cite{hu2017finding,zhang2020robust}, they only focus on the faces towards the camera, but crowd localization aims to locate all human heads (including extremely tiny heads) of various orientations in crowd. As a result, the prescion and recall of head detection are unsatisfactory for crowd analysis \cite{liu2018decidenet,gao2020cnn}. In this case, we address crowd localization by directly predicting the locations of head centers. \riii{Instead of obtaining the complete bounding boxes of human heads as detection methods do, our method aims only to find their locations and gives keypoint-based predictions (center of head if not overlapped or the shown part if overlapped). In this case, our method is more robust to overlapped heads because we do not have to find the complete heads and only part of the head is enough to make predictions. }

Most of the exsiting methods for crowd analysis are based on individual images, i.e., estimating the counting and localization results using a static image \cite{liu2019point,sam2020locate,liu2019recurrent,wen2019drone}. Because only the spatial information is utilized, these methods usually suffer from the issues of serious false positives and miss-detections in distant or blurry regions. Although some recent methods exploit temporal information \cite{xiong2017spatiotemporal,fang2019locality,liu2020estimating}, they only address crowd counting in videos without considering locations of human heads. We argue that it is crucial to incorporate temporal information of adjacent frames to discover the hard samples of human heads. To this end, we propose to address crowd localization in videos rather than images. Specifically, we aim to improve the localization results of the objective frame by also considering its adjacent frames.

\begin{table*}[tbp] 
		\caption{The summary of existing video datasets and VSCrowd for crowd counting and localization. Note that all datasets are collected by surveillance cameras expect for DroneCrowd.}
	\label{sets}
	\begin{center}
		\begin{tabular}{lcrrrrrcc}
			\hline
Datasets & Resolution & \#Frames & \#Annotations & Min & Max & Avg&\new{Trajectory} &\new{BBox} \\ \hline
UCSD \cite{chan2008privacy} & $238\times158$ & 2,000 & 49,885 & 11 & 46 & 24.9 \\
Mall \cite{change2013semi} & $640\times 480$ & 2,000 & 62,316 & 13 & 53 & 31.2 \\
WorldExpo \cite{zhang2015cross} & $720\times 576$ & 3,980 & 199,923 & 1 & 253 & 50.2 \\
Venice \cite{liu2019context}& $1280\times 720$ & 167 & 35,902 & 86 & 421 & 215.0 \\
FDST \cite{fang2019locality} & $1920\times 1080$ & 15,000 & 394,081 & 9 & 57 & 26.7 \\
DroneCrowd \cite{wen2019drone} & $1920\times 1080$ & 33,600 & 4,864,280 & 25 & 455 & 144.8 &\checkmark&\\ \hline
VSCrowd (Ours) & $1920\times 1080$ & 62,938 & 2,344,276 & 1 & 296 & 37.2&\checkmark&\checkmark \\ \hline
\end{tabular}
\end{center}
\end{table*}

Recently, mainstream crowd analysis methods usually rely on convolutional neural networks, the training of which requires large-scale annotated data. However, existing crowd datasets are either focused on single-image analysis or of small scale. The commonly-used video benchmarks for crowd counting and localization are summarized in TABLE \ref{sets}. We can see that UCSD \cite{chan2008privacy}, Mall \cite{change2013semi}, WorldExpo \cite{zhang2015cross}, Venice \cite{liu2019context}, and FDST \cite{fang2019locality} consist of small numbers of frames and annotations, which limits the generalizability of deep models. As for the recent DroneCrowd dataset \cite{wen2019drone}, it is captured by a drone-view camera. Although wider scenes are covered in DroneCrowd, the details of individuals are not retained. Apart from the datasets in TABLE \ref{sets}, there exist other video datasets recording pedestrians such as CrowdFlow \cite{schroder2018optical}, DukeMTMC \cite{ristani2016performance}, and PETS2009 \cite{ferryman2009pets2009}, but they focus on different tasks and cannot be directly applied to mainstream crowd analysis. For example, CrowdFlow is dataset of optical flow of moving individuals, which contains HD videos synthesized by Unreal Engine; DukeMTMC is a large-scale high-quality surveillance dataset with bounding box annotations and widely used for tracking, re-identification, and facial recognition, and it cannot be exploited for crowd analysis because the people in it are insufficient.
% which is less practical compared to surveillance in terms of public safety.
Hence, a representative surveillance-view benchmark is strongly desired for video crowd localization. \new{To facilitate research on this task, we create a large-scale video dataset named \textbf{VSCrowd}. The proposed dataset contains 62,938 high-resolution surveillance-view frames with 2,344,276 annotated pedestrians. Besides dot annotations for head centers, we also provide valuable bounding boxes and trajectories for each instance. It is worth noting that our video frames are captured in various scenes (e.g., malls, streets, scenic spots), as shown in Fig. \ref{samples}. In summary, VSCrowd is a significant contribution in terms of scalability, diversity, difficulty, and application.}

Intuitively, spatial-temporal dependencies are crucial for video crowd localization. Recently, the attention mechanism has achieved great success in various fields due to its strong capacity of capturing global dependencies \cite{devlin2018bert,carion2020end,dosovitskiy2020image,han2020survey,li2022video}. Inspired by this, we also employ the attention mechanism to model the correspondences among spatial-temporal features. However, the conventional attention mechanism (i.e., full attention) has two limitations: 1) The quadratic time and memory complexity with respect to sequence length seriously restricts the model from processing long sequences \cite{tay2020efficient,zhu2020deformable}; 2) Unnecessary and redundant information may be involved when computing the attention \cite{zaheer2020big}. To overcome these drawbacks, we propose Gaussian Neighborhood Attention (\textbf{GNA}) which samples a fixed number of keys for a query from a Gaussian distributed neighborhood centered at the position of the query. Different from full attention, GNA can capture long-range dependencies while maintaining the spatial topological structure of data. \new{Moreover, we equip GNA with a multi-focus mechanism which samples and aggregates the attentive features of various Gaussian ranges to handle the scale variation of head size caused by perspective effects. We have also theoretically proved that the proposed multi-focus attention mechanism can capture long-range dependencies within data.}

Based on the multi-focus GNA, we develop a unified framework for video crowd localization, called \textbf{GNANet}. It fully learns the spatial-temporal correlations to locate pedestrians in the objective frames (i.e., the middle frames) of crowd video clips. Specifically, GNANet first extracts semantic features separately from each frame in a video clip and then aggregate their features to model the scene context of the whole clip. By regarding the computed scene context as the query, a context cross-attention module captures the spatial-temporal information of videos with two sources of keys, i.e., \textit{feature of the objective frame} and \textit{features of all frames}. In particular, an objective-frame cross-attention is designed to capture the spatial dependencies between pixels of human heads within the objective frame, while a temporal cross-attention is applied for correlations between pixels in adjacent frames. By this means, the temporal information is aggregated into the objective frame for better localization. Finally, the attention outputs and the feature of the objective frame are combined to predict the localization map.

The contributions of this paper are summarized as follows:
\begin{itemize}
	\item A large-scale surveillance video dataset is created for crowd localization. This dataset consists of 634 videos (62,938 frames) and 2,344,276 annotations, with a comprehensive coverage of various crowd scenes.
	\item Multi-focus Gaussian neighborhood attention (GNA) is proposed to capture local dependencies and long-range dependencies with feature maps at various receptive field sizes for the scale modeling of human heads.  
	\item We develop a GNA-based network for video crowd localization, in which scene modeling and context cross-attention are utilized to retrain clip information and to model spatial-temporal correlations respectively. The proposed network achieves the state-of-the-art localization performance and comparable counting accuracy on several datasets.
\end{itemize}

The rest of this paper is organized as follows. We review some related works on crowd analysis and the attention mechanism Section \ref{rw}. A large-scale surveillance-view video crowd dataset (VSCrowd) is proposed in Section \ref{sd}. Then, the novel framework for video crowd localization is elaborated in Section \ref{pm}. To prove the effectiveness of our method, we conduct extensive experiments in Section \ref{exp}, including comparisons with existing methods, ablation study and further analysis. At last, we make some conclusions in Section \ref{con}.

\section{Related Work}
\label{rw}

\subsection{Crowd Counting and Localization}

\textbf{1) Single-image Crowd Counting.} Previous works on crowd analysis focus on crowd counting, i.e., estimating the total number of people in the crowd image \cite{huang2017body,liu2018crowd,li2018csrnet,tian2019padnet,bai2020adaptive,yan2019perspective,liu2019crowd,wan2021fine,liu2020crowd,mo2020background,cheng2021decoupled,yang2020embedding}. Numerous single-image crowd counting methods were proposed in recent years. For the first time, crowd counting was formulated as density map regression in \cite{zhang2016single}. In addition, considering the variations in head size due to perspective effect or image resolution, a multi-column convolutional neural network (MCNN) architecture was developed to map the crowd image to its crowd density map. Different to multi-scale architectures, CSRNet \cite{li2018csrnet} consists of a convolutional neural network (CNN) for feature extraction and a dilated CNN for density map regression. By replacing pooling operations with dilated kernels, larger reception fields are delivered. Inspired by the dilated convolution in CSRNet, the adaptive dilated convolution was proposed and leveraged for crowd counting in \cite{bai2020adaptive}. In addition, self-correction (SC) supervision was presented to iteratively correct the annotations and optimize the model considering the whole and the individuals. To deal with the variations of head size caused by the perspective effect, PGCNet \cite{yan2019perspective} uses perspective information to guide the spatially variant smoothing of feature maps, where the perspective information is estimated by a specifically designed branch. Context-aware crowd counting was proposed in \cite{liu2019context}, where features over multiple receptive field sizes are extracted and the position-wise importance of the features are learned. Different from the methods based on density map estimation, Bayesian loss \cite{ma2019bayesian} uses the point annotations to constructs a density contribution probability model. Experiments show that a standard backbone network trained by Bayesian loss outperforms that trained by the baseline loss to a large extent. The methods for pure crowd counting often generate a density map, the summation of which is the count of people in the image. Although this type of method cannot provide exact locations of individuals, they are inspiring in terms of network structure: like CSRNet, we use the encoder-decoder-style framework to extract semantic representations and predict the localization map.

\textbf{2) Single-image Crowd Localization.} Despite that state-of-the-art crowd counting methods can estimate the number of people in images with low error, they cannot predict the exact locations of heads. To overcome such limitation of crowd counting, crowd localization is also considered in recent works, and some multi-task methods for both counting and location are proposed \cite{liu2019recurrent,liu2019point,sam2020locate}. Similar to CSRNet, RAZNet \cite{liu2019recurrent} first extracts features from the crowd images, then predicts the density map and the localization map by two branches, respectively. In addition, to deal with high density region, a detection-and-zooming strategy is designed in RAZNet. Idrees {et al.} \cite{idrees2018composition} propose to simultaneously address crowd counting, density map estimation and localization in a unified framework consisting of multiple branches, each of which is responsible for generating a density map with specific degrees of sharpness. 
% When the sharpness approaches the infinity, the density map of the crowd image approximates the localization map that consists of binary values indicating the centers of heads. 
D2C \cite{cheng2021decoupled} decomposes density map estimation into two stages, probability map regression and count map regression. Besides, a peak point detection algorithm is proposed to localize the head centers under the guidance of local counts. 
% Despite that D2C can achieve crowd localization, the process is fixed and involves several manually defined parameters. 
A topological constraint is proposed in \cite{abousamra2021localization} to enable the model to learn the spatial arrangement of predicted head centers. Specifically, based on the theory of persistent homology, a persistence loss is defined to compare the topology of the ground truth and the topographic landscape of the likelihood map. A generalized loss function is presented in \cite{wan2021generalized} to train the network for crowd counting. It has been proved that pixel-wise $L_2$ loss and Bayesian loss are special cases of the generalized loss. Since the predicted density is pushed toward the head centers, the localization map can be easily obtained by simple post-processing. P2PNet \cite{song2021rethinking} is developed to directly predict the head centers in an image. The Hungarian algorithm is leveraged to perform the one-to-one matching between the predicted points and the ground truth ones, and then the loss is computed based on the matched points. 
% Since there is the one-to-one matching in training process, the optimization of P2PNet may cost much more computational time.
 Some detection-based crowd counting and localization methods are also explored recently. PSDDN \cite{liu2019point} first initializes the pseudo ground truth bounding boxes to train the detector. Then an online updating scheme is designed to modify the pseudo annotations during training. Similarly, a sophisticated scheme is developed for the generation of pseudo ground truth bounding boxes in LSC-CNN \cite{sam2020locate}. Then a multi-column architecture (equipped with top-down feature modulation) is exploited in LSC-CNN to aggregate multi-scale feature maps and to make predictions at multiple resolutions. The detection-based methods can predict not only the extract locations of human heads, but also their scales. State-of-the-art methods typically model this task as tiny object detection, which requires the pseudo bounding boxes for training. Such pseudo bounding boxes are estimated according to the geometric distances between the annotated points, so the quality cannot be guaranteed. In this case, we propose to directly estimate the binary localization map from the frames.
% Considering that RAZNet directly addresses crowd localization without complicated pre-processing for annotations, we borrow the localization branch in RAZNet to predict the localization map.

\textbf{3) Video Crowd Counting.} To further improve the performance of crowd analysis, some researchers tend to focus on video crowd counting, i.e. counting the people using spatial-temporal information \cite{xiong2017spatiotemporal,fang2019locality,liu2020estimating,wen2019drone}. Bidirectional ConvLSTM is exploited to captures both spatial and long-range temporal dependencies in video sequences \cite{xiong2017spatiotemporal}. A locality-constrained spatial transformer network (LSTN) is proposed in \cite{fang2019locality}, where the LST module estimates the density map of the next frame using that of the current frame. Besides, a dataset for video crowd counting (FDST) is proposed in this work. In \cite{liu2020estimating}, the people flows are estimated and summed to obtain the crowd density, where conservation constraints are imposed for robustness. Liu {et al.} \cite{liu2020counting} propose to estimate the people flow between frames and predict the crowd densities based on the flow. Compared with single-image crowd counting, this approach can impose constraints encoding the temporal dependencies between adjacent frames. Similar to the two-branch architecture of RAZNet, STANet was proposed to perform crowd counting and localization using frame sequences in \cite{wen2019drone}. In STANet, multi-scale feature maps of consecutive frames are aggregated to predict the density map and localization map of the current frame. Although our work and STANet address similar issues, we have different motivations and novel contributions: a) The scenes that we deal with are far different. Specifically, DroneCrowd consists of videos captured from drones, so the scale of human is tiny but consistent. In contrast, the videos in VSCrowd are surveillance, so the variation of human heads is large due to the strong perspective effect. Compared to drone-view videos that record the global dynamic of a large area, surveillance provides much more local details of a crowded spot, which we would like to use to perform fine-grained video analysis. b) In STANet, the temporal information is exploited by simply combining the feature maps of adjacent frames. However, in our method, we try to build a more effective video understanding model and have designed Gaussian neighborhood attention to capture the spatial-temporal dependencies in a video clip. Furthermore, the multi-focus mechanism is proposed to explicitly address the issue of the scale variation. In addition, a large-scale drone-view video crowd dataset (DroneCrowd) is proposed in \cite{wen2019drone}. \new{Although the crowd density in DroneCrowd is higher than that in the proposed VSCrowd, our dataset is superior to DroneCrowd in four folds: 1) The videos in our dataset are surveillances in which the strong perspective effect causes significant scale variations of human heads. However, the size of the human objects in DroneCrowd captured by drone-mounted cameras is consistent in the same video. Therefore, VSCrowd is more challenging that DroneCrowd. 2) The diversity of the human objects in VSCrowd greater in terms of face orientation, pose, size, etc. But the objects in DroneCrowd are merely a few pixels captured from high above. The diversity of our dataset can improve the robustness of deep models effectively. 3) VSCrowd has more applications in public safety because our videos are real surveillance of the crowd at various scenes in real life. 4) Besides the head center, we also annotate a bounding box covering the head-and-shoulder area for each person, with ID labeled. Hence, besides crowd counting and localization, VSCrowd can also be applied to other tasks such as tiny face recognition, pedestrian re-identification, and object tracking. In summary, our work has significant differences to STANet in methodology and dataset collection.}

% Most video crowd analysis works focus on crowd counting, i.e., boosting the performance of counting by exploiting the temporal information. In our work, following the setting of video crowd counting, we define video crowd localization as locating all human heads in a frame by considering the video clip. But different to STANet that addresses drone-view crowd analysis, we focus on video crowd localization in surveillance.

\subsection{Attention Mechanism}

The attention mechanism was first proposed in the field of natural language processing (NLP) to align the input sequence and the output sequence \cite{bahdanau2014neural}. Working with the LSTM-based encoder-decoder architecture, it has revealed great power in capturing long-range dependencies in sequences. Furthermore, researchers proposed Transformer based on the multi-head attention mechanism, where the global dependencies are captured in different aspects \cite{vaswani2017attention}. The state-of-the-art language model was constructed based on Transformer \cite{brown2020language}.

Inspired by its great success in NLP, the attention mechanism has been introduced to the field of computer vision \cite{ji2020spatio,hu2018squeeze,wang2018non,ramachandran2019stand,du2017recurrent,chen2019three,li2021groupformer}. Squeeze-and-excitation (SE) block was proposed to model the dependencies between channels \cite{hu2018squeeze}. By learning the channel-wise attention, the performance of CNN can be significantly improved at slight additional computational cost. The non-local block \cite{wang2018non} was developed to capture the long-range dependencies between pixels, which can be plugged into classic CNNs as a general operation. Numerous computer vision tasks such as video classification and object detection are improved. Motivated by the great success in processing sequences, Transformer has been leveraged for computer visions tasks. DETR was proposed in \cite{carion2020end}, where object detection is modeled as a sequence generation process in a encoder-decoder architecture. Considering that the computation complexity of self-attention is quadratic to the length of the sequence, numerous sparse attention mechanisms are proposed. For example, deformable DETR was proposed in \cite{zhu2020deformable}, where a query only attends to a small set of keys sampled from its neighborhood. Stand-alone self-attention was proposed in \cite{ramachandran2019stand}, where a query only attends to its surrounding keys. Based on the stand-alone self-attention mechanism, a fully attention vision model was proposed by replacing all spatial convolutions in this work. Essentially, the proposed Gaussian neighborhood attention is also a sparse attention, but it combines the advantages of the local attention and the random attention to capture the long-range dependencies between pixels without losing the topology structure of the video data, while costing less computational resources. Furthermore, multi-focus GNA is developed to deal with the scale variation caused by the strong perspective effect in surveillance.

\subsection{Multi-Focus Technique}

The multi-focus technique has been explored in other computer vision tasks recently, such as classification and segmentation. For example, a spatial-and-magnification-based attention sampling strategy is proposed in \cite{jin2020foveation}, where the attention network predicts a spatial probability distribution of informative patches at different magnifications. Although this method tries to understand large-scale histopathology images in multiple resolutions, it is different from our method which explicitly models the pixel-wise dependencies of various ranges to aggregate spatial-temporal information into the query pixel. The foveation module is introduced in \cite{zhang2021joint} to, for each position of the down-sampled image, estimate the importance of patches of different sizes around the position. By aggregating various patches for each position, the segmentation performance for mega-pixel histology images is improved. 
This method extracts multi-focus semantic information by simply cropping different sizes of patches and estimating their importance (patch-wise attention). In contrast, our method exploits pixel-wise attention to capture the long-range dependencies between pixels within Gaussian neighborhoods with various scales.

\section{VSCrowd Dataset}
\label{sd}

\begin{figure}[tbp]
	\centering
	\includegraphics[width=\columnwidth]{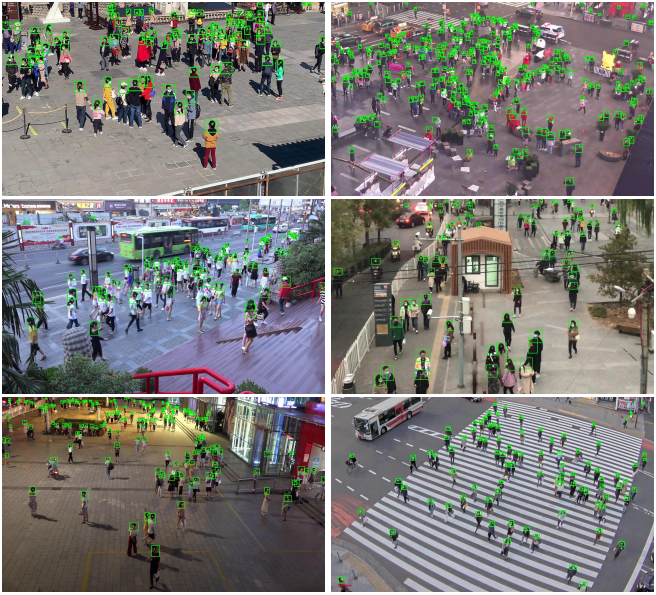}
	\caption{Some annotated samples from VSCrowd. Both head centers and head-shoulder boxes are provided. Various crowd scenes are included.}
	\label{samples}
\end{figure}

We create a large-scale video dataset named VSCrowd with a comprehensive coverage of various crowd scenes, such as squares, shopping malls, street crosses and campuses. The videos in VSCrowd are either collected from the Internet or captured by static cameras. The major difference between VSCrowd dataset and another large-scale video crowd dataset, DroneCrowd \cite{wen2019drone}, is that the videos in DroneCrowd are recorded by drone-view cameras, while VSCrowd focuses on surveillance-view. 
%In this case, VSCrowd is more suitable for the applications of public security.

Specifically, the videos in VSCrowd are collected at 25--30 frames per second (FPS) with the highest resolution of $1920\times 1080$. \riii{To reduce the temporal redundancy in videos, the collected videos are sub-sampled to 5 FPS. Similar pre-processing is also applied for DroneCrowd.}
Then we segment all videos into shots that are about 20 seconds. That is to say, each shot contains approximately 100 frames. Finally, 634 shots (62,938 frames in total) are collected. \new{The annotation of the proposed VSCrowd was carried out on SenseBee that is an annotation platform designed by SenseTime Group Ltd. Before annotation, we set strict protocols for the process, including the explicit definition of head centers and head-shoulder area, the way to deal with blurred/occluded region, and the solutions of several special cases (people in mirror/cars or on bicycles, babies in strollers, etc.). During annotation, several videos were assigned to each annotator, and the annotators were asked to provide the following ground truth for each frame: the head centers, the bounding boxes covering the head-should area, and the unique ID for each person. It is worth mentioning that we allow the annotators to reasonable estimate the locations of blurred people by investigate the adjacent frames. Hence, for each person in a video, we know its exact location at arbitrary time. It took over 30 well-trained annotators three months to complete the annotation of all videos. After annotation, extra workers are responsible for checking the correctness and quality of the labels. After checking and refinement, we ensure that the accuracy of the labels is more than 97\%. Finally, 2,344,276 head labels are obtained (i.e., about 3,698 per shot, 37 per frame). Fig. \ref{samples} shows some annotated frames from the proposed VSCrowd.}

% In terms of annotation, more than 25 expert annotators annotate all head centers and the head-shoulder boxes in all frames. The head-shoulder box covers the entire head and shoulder of a person to provide more details for detection. \new{The frame-wise head-shoulder boxes and the head centers are labeled by carefully examining the video clips. Specifically, when labeling a specific frame in a video, the workers are required to investigate its several adjacent frames to determine the exact locations of human heads. For the heads that are partially blurred/occluded or too far to see clearly, we allow the workers to estimate the locations by looking into the adjacent frames. For the heads that are completely occluded or outside the boundary of the video, they are ignored with no annotation. After annotation, extra workers are responsible for checking the correctness and quality of the labels. Finally, we ensure that the accuracy of the labels is more than 97\%.} After annotation, 2,344,276 head labels are obtained (i.e., about 3,698 per shot, 37 per frame). Fig. \ref{samples} shows some annotated frames from the proposed VSCrowd.

\begin{figure}
     \centering
     \begin{subfigure}[b]{0.8\columnwidth}
         \centering
         \includegraphics[width=\textwidth]{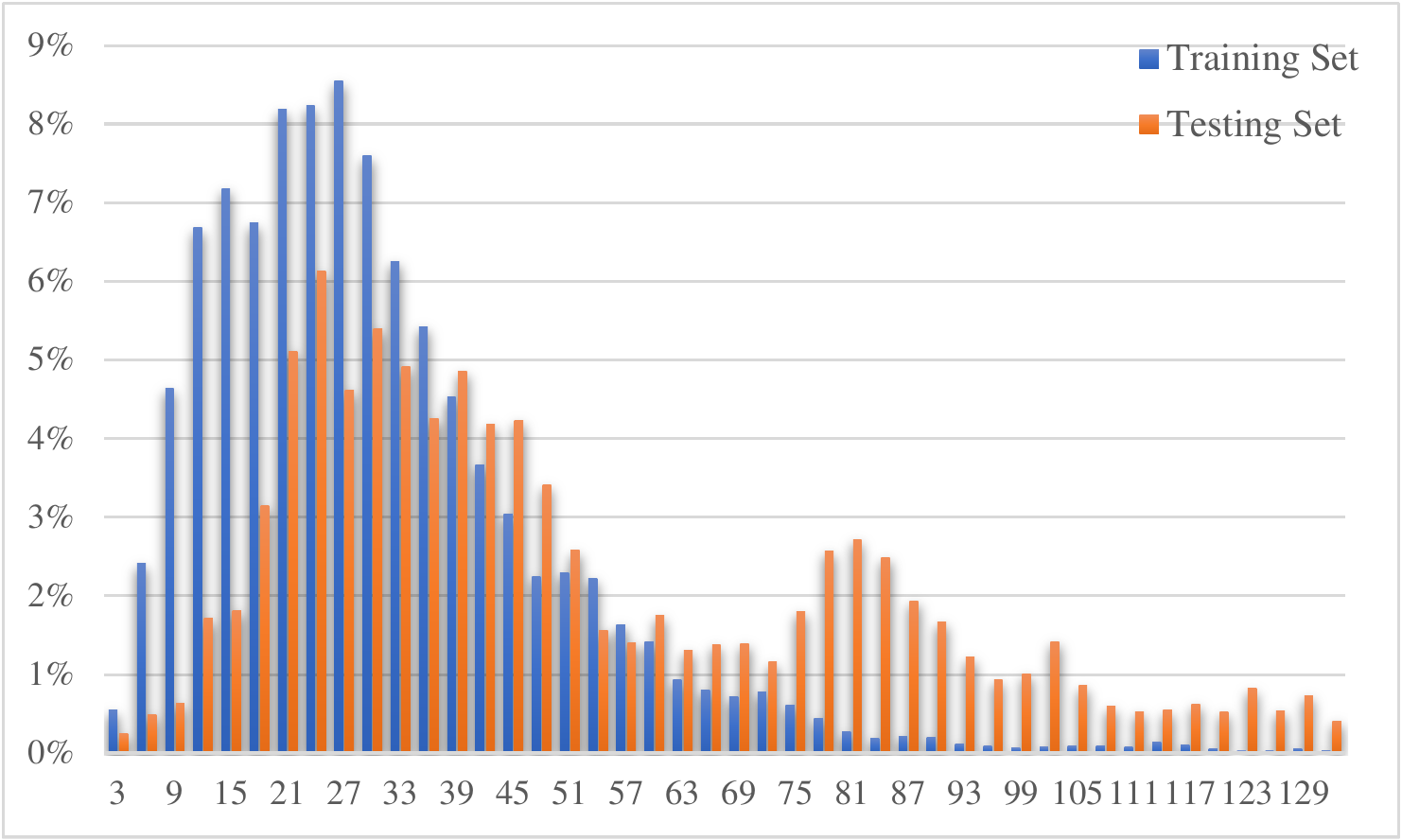}
         \caption{The distribution of the number of people.}
         \label{dist}
     \end{subfigure}
     \hfill
     \begin{subfigure}[b]{0.8\columnwidth}
         \centering
         \includegraphics[width=\textwidth]{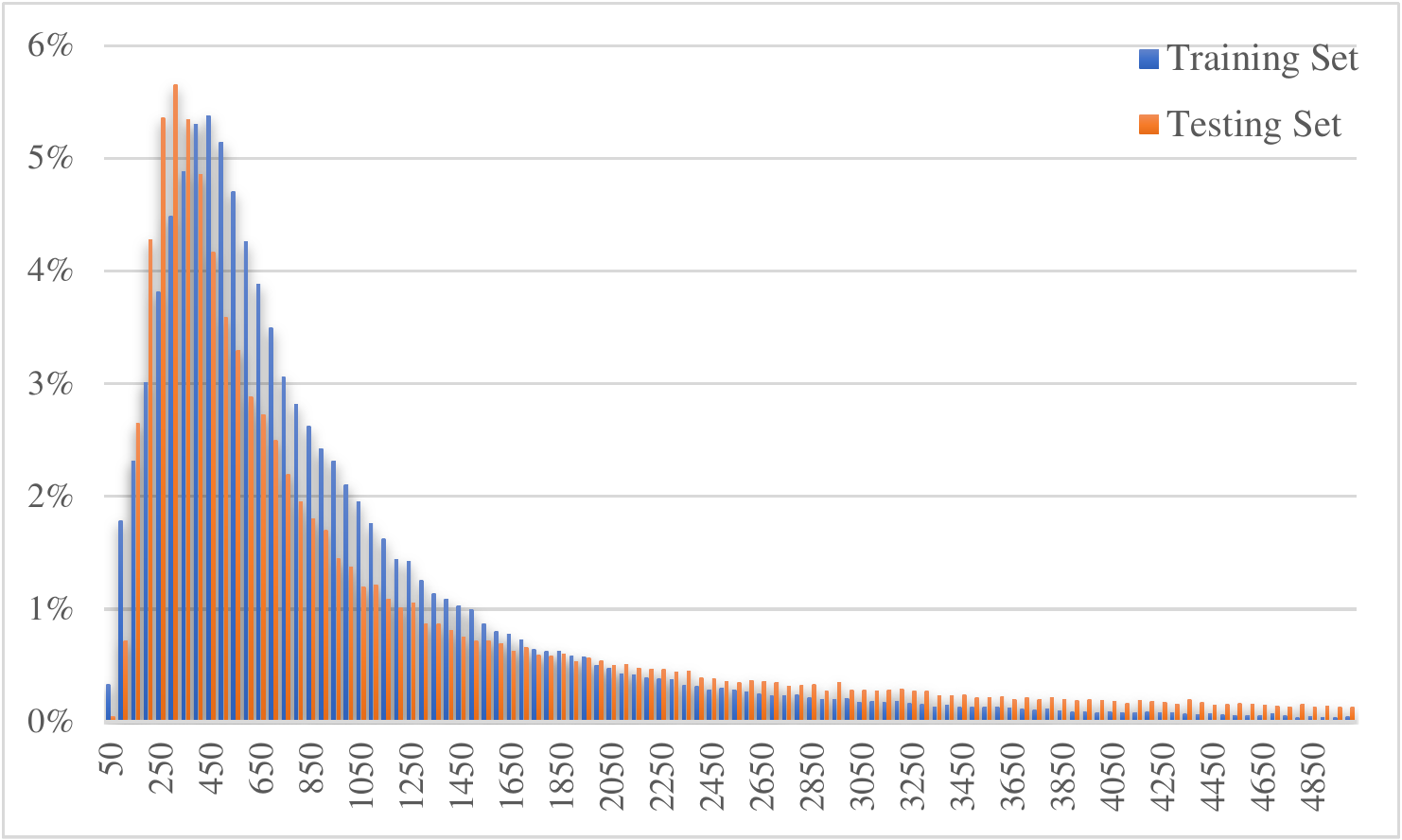}
         \caption{The distribution of the head size in pixels.}
         \label{adist}
     \end{subfigure}
        \caption{\new{Distributions of VSCrowd.}}
        \label{fig:three graphs}
\end{figure}

Regarding the constructions of the training set and the testing set, we split VSCrowd at the video level, i.e., the training set and the testing set contain different videos. Furthermore, to make our dataset more challenging, we make sure that the scenes in the training set are different from those in the testing set. The reason why we split VSCrowd in this manner is that we would like the model to learn scene-agnostic representations for crowd localization by training and testing it on videos with different scenes. In this case, a model that has better performance on our dataset means it attains great generalization ability and is more powerful when dealing with new scenes. Specifically, VSCrowd is divided into a training set (497 shots/78.4\%) and a testing set (137 shots/21.6\%), and there exists a large scene gap between these two parts. \new{We provide more statistics about the proposed VSCrowd, including the distribution of the number of people in each frame (Fig. \ref{dist}) and the distribution of the size of heads in pixels (Fig. \ref{adist}). As shown in the figure, the testing set contains more people than the training set. The distribution gap between the training set and the testing set makes our dataset more challenge. In terms of the distribution of the head size, it is long tailed in both the training set and the testing set. The distribution indicates there is a large variation in the head size in the proposed dataset, which is a distinct characteristic compared with other crowd datasets.}

% VSCrowd is divided into a training set (497 shots/78.4\%) and a testing set (137 shots/21.6\%). Note that the scenarios in the training set are different from those in the testing set. It means there exists a large distribution gap between these two parts. By this means, VSCrowd can better evaluate the generalizability of machine learning models.

\section{The Proposed Method}
\label{pm}

% \subsection{Problem Definition}

% Video crowd localization aims to locate all head centers in the keyframe from a given video clip. Given a video clip ${V}^i=\left\{\bm{f}^i_{j}\right\}_{j=0}^{m-1}$ ($\bm{f}^i_{j}\in\mathbb{R}^{H\times W\times 3}$, $m$ is the clip length, $i$ is the clip index), the middle frame $\bm{f}^i_c$ ($c:={\left\lfloor \frac{m}{2}\right\rfloor}$) is defined as the keyframe of this clip. The ultimate goal is to compute a probability map $\hat{\bm{L}}^i\in\mathbb{R}^{H\times W}$ for $\bm{f}^i_c$, where $\hat{\bm{L}}^i_{\bm{p}}$ gives the probability that the position $\bm{p}=(x,y)$ in $\bm{f}^i_c$ is a head center. 

In this section, we elaborate the proposed multi-focus Gaussian neighborhood attention (GNA) module and design a new structure for crowd localization based on multi-focus GNA.

\subsection{Multi-Focus Gaussian Neighborhood Attention}

\new{The attention mechanism is well exploited in computer vision \cite{chen2020pre,prangemeier2020attention}. Given a query set $\left\{\bm{q}_j\right\}$ and a key-value pair set $\left\{(\bm{k}_i,\bm{v}_i)\right\}$, where $\bm{q}_j,\bm{k}_i,\bm{v}_i \in \mathbb{R}^d$, the attention output for $\bm{q}_j$ is computed as follows,
\begin{align}
{\rm Attn}(\bm{q}_j)&=\sum_i{\rm softmax}_i( s(\bm{q}_j,\bm{k}_i))\bm{v}_i,\notag\\
&=\sum_i\frac{\exp\{s(\bm{q}_j,\bm{k}_i)\}}{\sum_{k}\exp\{s(\bm{q}_j,\bm{k}_k)\}}\cdot\bm{v}_i,
\end{align}
where $s(\bm{q}_j,\bm{k}_i)$ is the score function measuring the similarity between key $\bm{k}_i$ and query $\bm{q}_j$. Essentially, attention mechanism is the weighted average of the values based on the similarity between the query and the corresponding key. Scaled dot-product similarity \cite{vaswani2017attention} is well exploited and reveals great stability in recent works \cite{dosovitskiy2020image}. In this paper, we use scale dot-product as the score function, i.e.,  $s(\bm{q}_j,\bm{k}_i)=\frac{\bm{q}_j^{\rm T}\bm{k}_i}{\sqrt{d}}$. }

A major concern with conventional full attention is the quadratic time and memory complexity with respect to sequence length, which limits its applications in many settings. Due to the overwhelming complexity of full attention, many efficient variants are proposed in recent literature. For example, \cite{ramachandran2019stand,hu2019local} proposed local attention where each query can only be connected to keys within its fixed neighborhood, so the attention retains the topological structure in the data. However, local attention is not capable of capturing the dependency between a query and keys outside the fixed neighborhood. Random attention is exploited in \cite{zaheer2020big}, where each query attends to a small number of keys uniformly sampled from the whole set, so each query captures certain global information with linear computational cost. But redundancy are involved since two distant pixels are unrelated in general.

\begin{figure*}[tbp]
	\centering
	\includegraphics[width=\textwidth]{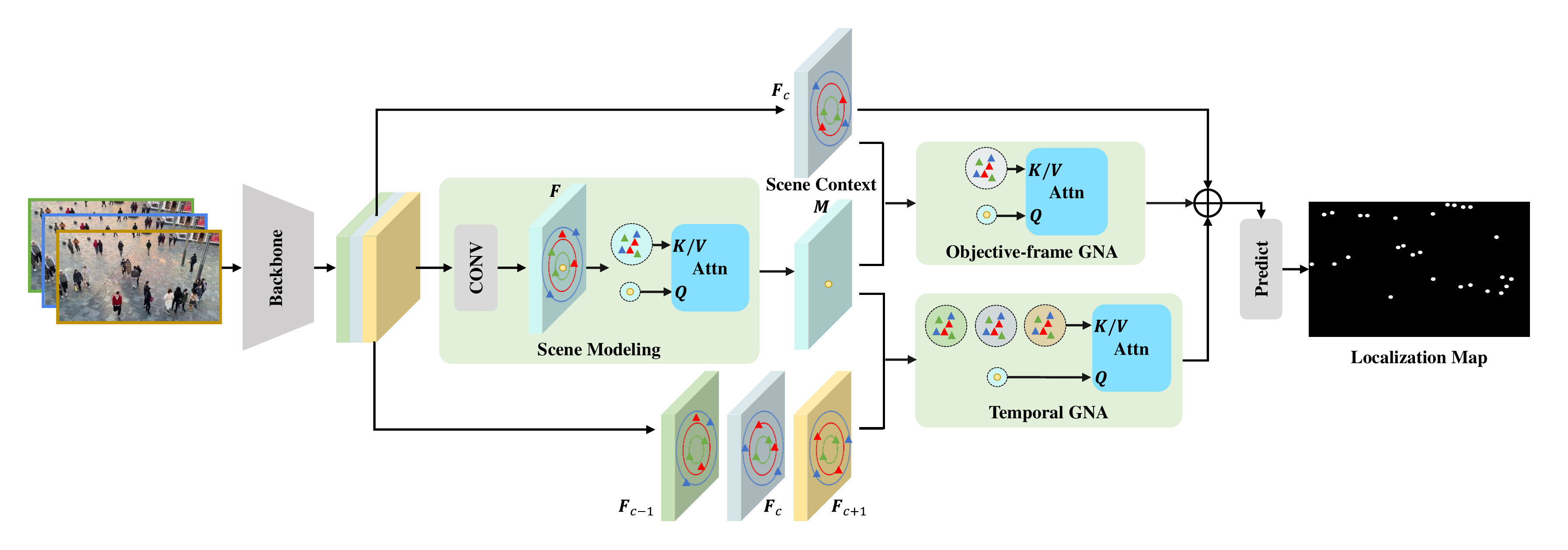}
	\caption{The overview of the proposed GNANet. Taking a video as input, GNANet predicts the localization map of the objective frame using scene modeling and context cross-attention.}
	\label{overview}
\end{figure*}

In this paper, we propose Gaussian neighborhood attention (GNA), where each query attends to keys sampled from its Gaussian random neighborhood. Essentially, the proposed Gaussian neighborhood attention is a special case of random neighborhood attention (RNA). In this case, we first explain the general RNA and then specify GNA based on RNA as follows. \new{Taking two-dimensional feature maps for example, assuming $\bm{Q},\bm{K},\bm{V}\in \mathbb{R}^{H\times W\times d}$ are queries, keys, and values respectively, for each position $\bm{x}$ on $\bm{Q}$, we formulate RNA in terms of expectations over the neighborhood sampling distribution as follows,
\begin{equation}
\label{123}
{\rm RNA}(\bm{Q}_{\bm{x}},N,\mathcal{P}_N)=\mathbb{E}_{(\bm{x}_1,\cdots,\bm{x}_N)\sim\mathcal{P}_N}\left[{\rm Attn}(\bm{Q}_{\bm{x}})\right],
\end{equation}
\begin{equation}
\label{456}
{\rm Attn}(\bm{Q}_{\bm{x}})=\sum_{i=1}^N{\rm softmax}_i\left(\frac{\bm{K}_{\bm{x}_i}^{\rm T}\bm{Q}_{\bm{x}}}{\sqrt{d}}\right)\bm{V}_{\bm{x}_i},
\end{equation}
where $N$ is the pre-defined number of sampled points for attention, and $\mathcal{P}_N$ is a joint distribution of $N$ independent and identical 2D distributions with mean $\bm{x}$ such as 2D Gaussian, Laplace, and Student’s \textit{t}. Therefore, $\mathcal{P}_N$ is a $2N$-D distribution with mean $\left(\bm{x},\cdots,\bm{x}\right)\in\mathbb{R}^{2N}$. By applying different distributions to $\mathcal{P}_N$, different RNA mechanisms are constructed. For instance, fully random attention is constructed by applying uniform distribution over the whole spatial range. In this paper, we propose to use 2D Gaussian distribution with pre-defined standard deviation $\gamma$ for RNA, and the Gaussian neighborhood attention is specified.}
% The reasons why we choose to use Gaussian distribution for sampling instead of other heavy-tail distributions is: The standard deviation we set for Gaussian distribution is large (such as 5 pixels), so that the sampled points, theoretically, contain information within a relatively large region of 706 pixels. Such a region in a feature map contains enough semantic information for the attention of the centered query, so those heavy-tail distributions are unnecessary.
% b) According to the central limit theorem (CLT), the distribution of the average of random variables tends toward a Gaussian distribution. Therefore, it is more reasonable to choose Gaussian distribution when there is no prior information of the crowd images.

The pre-defined standard deviation $\gamma$ in GNA is crucial, as it controls the range of the region in which the information is aggregated. Compared with the uniform distribution over a fixed circular neighborhood, the Gaussian sampling around the query point has the following advantage. In GNA, most sampled points fall into the region within 3 standard deviations $3\gamma$ around the mean, and the closer to the query, the denser the sampled points, which is preferred since, in general, the pixels closer to the query have more correlations with the query. However, if we sample points uniformly within a circular neighborhood of radius $3\gamma$, more sampled keys can appear far away from the center point, introduce more less correlated keys which are less informative for the query into the computation of attention.

A distinct characteristic of surveillance is the strong perspective effect which would cause scale variation of objects. It is a major challenge in detection tasks as mentioned in previous works, and it may lead to multiple detections for one object (low mAP) or miss-detections (low mAR). To overcome the severe scale variation of human heads in VSCrowd, multi-focus GNA (MF-GNA) is utilized in our model. Specifically, the standard deviation $\gamma$ is set to different values and the output of MF-GNA is computed by average. \new{Such practice can also improve the ability of the model in capturing long-range dependencies, which is explained as follows. Formally, the definition of the proposed MF-GNA is expressed as follows,
\begin{equation}
{\rm MFGNA}=\frac{1}{F}\sum_{f=1}^F\mathbb{E}_{\bm{z}\sim\mathcal{N}_N^{\gamma_f}}\left[{\rm Attn}(\bm{Q}_{\bm{x}})\right],
\end{equation}
where $\bm{z}:=(\bm{x}_1,\cdots,\bm{x}_N)$, $F$ is the pre-defined number of focuses, and $\mathcal{N}_N^{\gamma_f}$ is a $2N$-D Gaussian distribution with mean $\left(\bm{x},\cdots,\bm{x}\right)\in\mathbb{R}^{2N}$ and standard deviation $\gamma_f$ at each dimension. It can be derived that
\begin{align*}
{\rm MFGNA}&=\frac{1}{F}\sum_{f=1}^F\mathbb{E}_{\bm{z}\sim\mathcal{N}_N^{\gamma_f}}\left[{\rm Attn}(\bm{Q}_{\bm{x}})\right]\\
&=\frac{1}{F}\sum_{f=1}^F\int p_{\mathcal{N}_N^{\gamma_f}}(\bm{z}){\rm Attn}(\bm{Q}_{\bm{x}})\mathrm{d}\bm{z}\\
&=\int\left[\sum_{f=1}^F\frac{1}{F} p_{\mathcal{N}_N^{\gamma_f}}(\bm{z})\right]{\rm Attn}(\bm{Q}_{\bm{x}})\mathrm{d}\bm{z}.
\end{align*}
Let $\mathcal{M}$ denote the uniform mixture of $\left\{\mathcal{N}_N^{\gamma_f}\right\}_{f=1}^F$,
\begin{equation}
p_{\mathcal{M}}(\bm{z})=\sum_{f=1}^F\frac{1}{F} p_{\mathcal{N}_N^{\gamma_f}}(\bm{z}).
\end{equation}
Then,
\begin{align*}
{\rm MFGNA}&=\int\left[\sum_{f=1}^F\frac{1}{F} p_{\mathcal{N}_N^{\gamma_f}}(\bm{z})\right]{\rm Attn}(\bm{Q}_{\bm{x}})\mathrm{d}\bm{z}\\
&=\int p_{\mathcal{M}}(\bm{z}){\rm Attn}(\bm{Q}_{\bm{x}})\mathrm{d}\bm{z}\\
&=\mathbb{E}_{\bm{z}\sim\mathcal{M}}\left[{\rm Attn}(\bm{Q}_{\bm{x}})\right]={\rm RNA}(\bm{Q}_{\bm{x}},N,\mathcal{M}).
\end{align*}
Therefore, the proposed multi-focus GNA is equivalent to the random neighborhood attention with the uniform mixture Gaussian. Furthermore, according to \cite{wainwright1999scale}, we can assume there exists a long-tailed distribution $\mathcal{H}$ such that
\begin{equation}
p_{\mathcal{H}}(\bm{z})\approx p_{\mathcal{M}}(\bm{z}).
\end{equation}
Then the proposed multi-focus GNA can be further approximated as follows,
\begin{align*}
{\rm MFGNA}&=\int p_{\mathcal{M}}(\bm{z}){\rm Attn}(\bm{Q}_{\bm{x}})\mathrm{d}\bm{z}\approx\int p_{\mathcal{H}}(\bm{z}){\rm Attn}(\bm{Q}_{\bm{x}})\mathrm{d}\bm{z}\\
&=\mathbb{E}_{\bm{z}\sim\mathcal{H}}\left[{\rm Attn}(\bm{Q}_{\bm{x}})\right]={\rm RNA}(\bm{Q}_{\bm{x}},N,\mathcal{H}).
\end{align*}
In conclusion, the proposed multi-focus GNA can also be viewed as a random neighborhood attention with a long-tailed distribution. Therefore, the multi-focus mechanism can improve the ability of the model in capturing long-range dependencies between pixels.} Equipped with multi-focus mechanism, GNA achieves a good balance between mAP and mAR.
% Figure \ref{overview} (\textit{Right}) illustrates 3-focus GNA operation between feature maps (query map and key map). 
For simplicity, in the rest of this paper, GNA refers to multi-focus GNA unless otherwise specified.

% For instance, if we use Gaussian distribution to sample the points, RNA is specified as the proposed Gaussian neighborhood attention (GNA).
% Taking two dimensional feature maps for example, assuming $\bm{Q},\bm{K},\bm{V}\in \mathbb{R}^{H\times W\times d}$ are queries, keys, and values respectively, for each position $\bm{p}=(x,y)$ on $\bm{Q}$, a two dimensional Gaussian distribution $\mathcal{N}(\bm{p},\gamma^2\bm{I}_2)$ is created, where $\gamma$ is the standard deviation controlling the attention range. Then a fixed number of points are sampled from $\mathcal{N}(\bm{p},\gamma\bm{I}_2)$, denoted as $\left\{(x_i,y_i)\right\}_{i=1}^P$ ($P\ll HW $). The GNA output for $\bm{Q}_{xy}\in \mathbb{R}^d$ is computed as follows,
% \begin{equation}
% 	{\rm GNA}(\bm{Q}_{xy})=\sum_i{\rm softmax}_i\left(\frac{\bm{K}_{x_iy_i}^{\rm T}\bm{Q}_{xy}}{\sqrt{d}}\right)\bm{V}_{x_iy_i}.
% \end{equation}

With the advantages of both local attention and random attention, the proposed GNA not only maintains the topological structure of the original data, but also captures long-range dependencies with linear computational cost. Theoretical analysis on the proposed GNA with respect to other sparse attention mechanisms can be found in Appendix \ref{ta}. Note that the sampling operation is a leaf node in the computational graph, so GNA is differentiable with respect to the input feature maps ($\bm{Q},\bm{K},\bm{V}$). \new{Due to Gaussian sampling in GNA, there exists randomness in the training and testing of the model. The randomness in training acts like dropout in typical neural networks, which can improve the robustness of GNANet. However, the randomness in testing is unwanted. To increase the stability of our model in testing, we approximate the expectation using Monte Carlo integration,
\begin{align*}
{\rm GNA}_{test}&=\mathbb{E}_{(\bm{x}_1,\cdots,\bm{x}_N)\sim\mathcal{N}_N^{\gamma}}\left[{\rm Attn}(\bm{Q}_{\bm{x}})\right]\\
&\approx \frac{1}{T}\sum_{t=1}^T \sum_{i=1}^N{\rm softmax}_i\left(\frac{\bm{K}_{\bm{x}^t_i}^{\rm T}\bm{Q}_{\bm{x}}}{\sqrt{d}}\right)\bm{V}_{\bm{x}^t_i},
\end{align*}
where $\left\{(\bm{x}^t_1,\cdots,\bm{x}^t_N)\right\}_{t=1}^T$ is the $T$ sets of sampling results of $(\bm{x}_1,\cdots,\bm{x}_N)$. In practice, we compute GNA multiple times and the average is taken as the final output of the GNA module during the testing phase.}

% we leverage the multi-sample mechanism, where GNA is computed multiple times and the average is taken as the final output of the GNA module. The reason why we do not increase the number of sampled points is that sampling too many points in GNA would bring great computational memory costs in practice. Additionally, averaging the results of multiple sample times essentially estimates the expectation of GNA over the sampled distribution.

%	The key pixels are sampled 3 times in GNA when testing to achieve the balance between efficiency and stability. Note that we do not use multi-sample mechanism in training because random sampling increases the robustness of the model.

\subsection{Video Crowd Localization via GNA }

In this subsection, we propose a new network structure based on multi-focus Gaussian neighborhood attention, called GNANet, for video crowd localization. \riii{Note that, to reduce the temporal redundancy between adjacent frames, we first sub-sample the videos to a low frame rate before performing crowd localization.
The overview of GNANet is shown in Fig. \ref{overview}. It takes a frame sequence (after sub-sampling) as input and predicts the localization map of the objective frame that is defined as the center frame.} GNANet consists of four modules: backbone, scene modeling module, context cross-attention module and localization prediction module. Each of them is elaborated as follows.

\textbf{1) Backbone.} The backbone takes each frame in a clip as input independently and extracts semantic features from each frame. Considering the fact that VGGNet \cite{simonyan2014very} shows great power in crowd analysis \cite{liu2019recurrent,li2018csrnet,luo2020hybrid}, the pre-trained VGGNet is adopted as the backbone. Specifically, we use the first 13 layers in VGG16 to extract informative features from frames. Given a video clip consisting of $m$ frames ${V}^i=\left\{\bm{f}^i_{j}\right\}_{j=0}^{m-1}$, where $\bm{f}^i_{j}\in\mathbb{R}^{H\times W\times 3}$ and $i$ is the clip index, the backbone extracts features from each frame. The feature maps of the clip ${V}^i$ are denoted as $\left\{\bm{F}^i_{j}\right\}_{j=0}^{m-1}$, where $\bm{F}^i_{j}\in \mathbb{R}^{\frac{H}{8}\times \frac{W}{8}\times 512}$. Note that the middle frame $\bm{f}^i_c$ ($c:={\left\lfloor \frac{m}{2}\right\rfloor}$) is defined as the objective frame of this clip. 
For simplicity, the clip index $i$ is omitted in the rest of this subsection.

\textbf{2) Scene Modeling.} Scene modeling is of great importance for crowd analysis, especially for surveillance where the camera is static. In this work, we compute scene context using the proposed GNA. Concretely, the features of all frames in a clip $\bm{F}_{j}\in \mathbb{R}^{\frac{H}{8}\times \frac{W}{8}\times 512}$ are aggregated by convolution (with input channel $512m$, output channel 512, and kernel size $(3,3)$) and ReLU. The aggregated feature of a clip is denoted as $\bm{F}\in\mathbb{R}^{\frac{H}{8}\times \frac{W}{8}\times 512}$. To capture the long-range dependency in the video clip, Gaussian neighborhood self-attention (GNA) is conducted on $\bm{F}$. Specifically, $\bm{Q},\bm{K},\bm{V}$ for GNA in Eq. \ref{123} and Eq. \ref{456} are all defined as $\bm{F}$. The attention output and the aggregated feature $\bm{F}$ are concatenated along channel dimension and sent into a convolutional layer to obtain the final scene context $\bm{M}\in\mathbb{R}^{\frac{H}{8}\times \frac{W}{8}\times 512}$. \riii{Since we apply the proposed GNA to aggregate the information from all frames, the obtained scene context $\bm{M}$ contains the important pixel-wise semantic dependencies in the frame sequence.}

\textbf{3) Context Cross-attention.} After obtaining the scene context $\bm{M}$ of a clip, the network retrieves information from the whole clip and the objective frame respectively. To retrieve information from the whole clip, we sample spatial locations on all frames in the clip for each position on $\bm{M}$. Specifically, for each spatial location on $\bm{M}$, a fixed number of points are sampled from each frame in a clip. The union of the sampled points of all frames is taken as the key-value index set. Then we compute Temporal GNA by treating $\bm{M}$ as query and sampled features from all frames as key/value. To retrieve information from the objective frame, we directly compute Objective-frame GNA by treating $\bm{M}$ as query and the sampled features from the objective frame as key/value, i.e., $\bm{K}=\bm{V}=\bm{F}_c$, $\bm{Q}=\bm{M}$. Finally, the element-wise sum of the two attention outputs and the feature of the objective frame $\bm{F}_c$ is sent into the prediction module. \riii{Note that although we aim to predict the localization result of the objective frame, we do not expect a perfect feature of the objective frame, but, instead, we exploit the spatial-temporal feature of the short video clip containing the object frame. The obtained spatial-temporal feature is expected to be aware of which the object frame is, rather than focusing only on it.}

\textbf{4) Localization Prediction.} The input of the localization prediction module is a 512D feature map containing the spatial-temporal information of the clip. Note that the spatial size of the feature map is $1/8$ of that of the original frame, so the localization prediction module is required to up-sample the feature map to a single-channel localization map with the same spatial size as the original frame. In this work, we follow the design of the localization branch in RAZNet \cite{liu2019recurrent} to obtain the localization map. Specifically, three convolution layers (with output channel 512, kernel size (3,3), dilation (2,2)) are first applied to further extract the semantic features. Then, the feature map is further processed by three deconvolution layers (with kernel size (3,3), stride (2,2), output channel 256/128/64), each of which is followed by a convolution (kernel size (3,3), dilation (2,2)) without changing the number of channels. Note that all convolution/ deconvolution layers are followed by ReLU activation. Since each deconvolution layer doubles the spatial size of the feature map, the output is a 64D feature map of the same size as the original image. Finally, an $1\times 1$ convolution (with sigmoid activation) converts the feature map into a single-channel localization map indicating the probability that a pixel is a head center.

\subsection{Implementation Details}
The training of GNANet follows the standard procedure of supervised learning. We denote the training set as $\left\{\left({V}^i,\bm{L}^i\right)\right\}_{i=1}^M$, where $\bm{L}^i\in\{0,1\}^{H\times W}$ is the ground truth binary map for the objective frame of the cilp ${V}^i$, obtained by processing the head annotations as in \cite{liu2019recurrent}. Given GNANet (denoted as $G_{\bm{\theta}}$ parameterized by $\bm{\theta}$),
% Given GNANet ($G_{\bm{\theta}}$ parameterized by $\bm{\theta}$) and the training set $\left\{\left({V}^i,\bm{L}^i\right)\right\}_{i=1}^N$ ($\bm{L}^i\in\{0,1\}^{H\times W}$ is the ground truth binary map for the keyframe of the cilp ${V}^i$, which is obtained by processing the head annotations as in \cite{liu2019recurrent}), 
the objective is to solve the problem as follows,
\begin{equation}
	\bm{\theta}^*= \mathop{\arg\min}_{\bm{\theta}}\frac{1}{M}\sum_{i=1}^M \mathcal{L}\left(G_{\bm{\theta}}({V}^i),\bm{L}^i\right),
\end{equation} 
where $\mathcal{L}$ is the loss function. Specifically, we use weighted binary cross entropy loss where the positive weight is set to 50 to compensate the heavy imbalance between the positives and negatives in the localization map.

% \noindent \textbf{Implementation details.} 
% The direct output of GNANet is a probability map where the value at each position indicates the probability that the position is a head center. In this case, it needs further processing to obtain the precise head locations. 
% Intuitively, the local peaks of the probability map can be regarded as the head centers. 
For post-processing, we follow the implementation in  \cite{liu2019recurrent,wen2019drone} to compute the local peaks of the localization map and predict the exact locations of human heads. In consideration of performance and GPU memory limitation, we set the length of video clips $m=3$. As for the proposed multi-focus GNA, the standard deviation $\gamma$ of Gaussian distribution is determined as 3, 5 and 10. In most crowd datasets, the size of a tiny human head can only be several pixels (even one pixel in extreme cases) on the feature map. For the relatively large heads, their sizes would not exceed 30 pixels on the feature map. In terms of the smallest standard deviation ($\gamma=3$) we choose, the sampled keys are almost in the circle of radius 9 centered at the query pixel, while for the largest one ($\gamma=10$), the radius is 30. The Gaussian neighborhood is large enough to cover the region of a human head. In addition, 32 points are sampled as the key-value indexes for each focus. More analysis on the hyper-parameters in GNA is conducted in Section \ref{fura}. In terms of training, our model is trained for 30 epochs using mini-batch stochastic gradient descent at a fixed learning rate $2\times 10^{-6}$. The batch size is set to 16. All experiments are conducted on NVIDIA V100 (32 GB) using PyTorch.

\section{Experiments}
\label{exp}
\begin{figure*}[tbp]
	\centering
	\includegraphics[width=\textwidth]{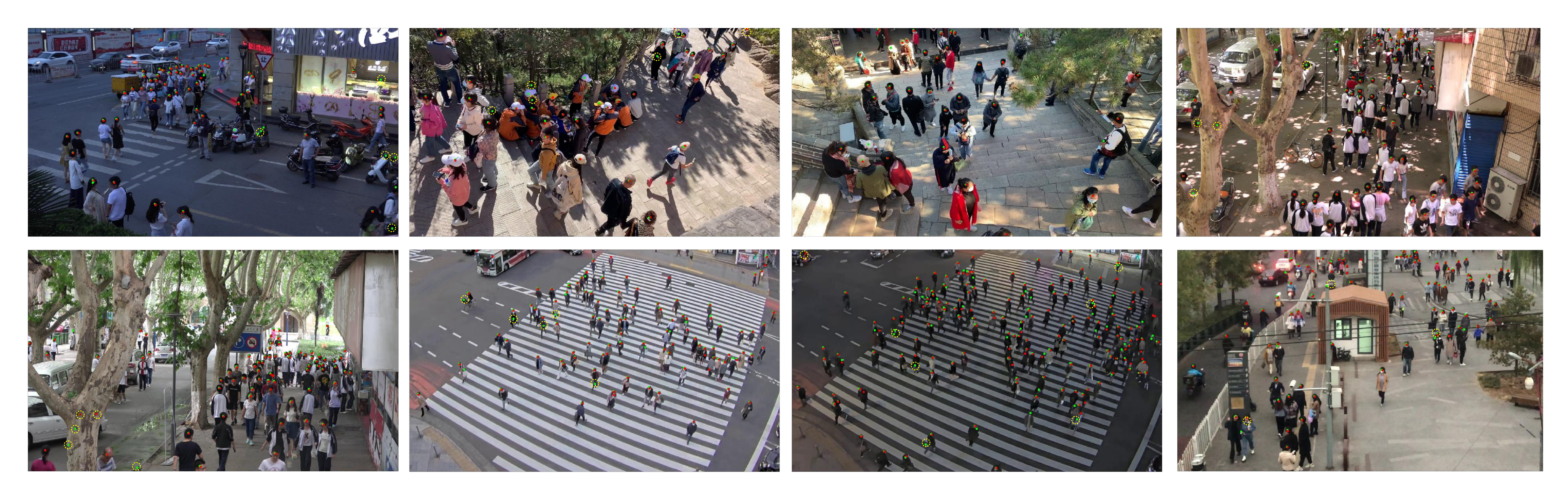}
	\caption{Some localization results of VSCrowd predicted by RAZNet (\textcolor[RGB]{0,255,0}{green} points) and the proposed GNANet (\textcolor[RGB]{255,0,0}{red} points). The false predictions are highlighted by \textcolor[RGB]{255,255,0}{yellow} dashed circles. Better viewing by zooming in.
% 	The \textcolor[RGB]{0,255,0}{green} points are the ground truth head centers, while the \textcolor[RGB]{255,0,0}{red} ones are the predictions. Better viewing by zooming in.
}
	\label{outputs}
\end{figure*}

In this section, we first explain the evaluation metrics for localization and couning. Then we compare the proposed GNANet with previous localization methods and counting methods on VSCrowd, FDST \cite{fang2019locality}, Venice \cite{liu2019context}, DroneCrowd \cite{wen2019drone}, ShanghaiTech \cite{zhang2016single}, UCF-QNRF \cite{idrees2018composition}, and NWPU \cite{wang2020nwpu}. Finally, we conduct ablation study and further analysis on GNANet to verify the effectiveness of the proposed modules. 

\subsection{Evaluation Metrics}

In this work, we use the evaluation protocol of key point detection in MS-COCO \cite{lin2014microsoft} to evaluate the localization methods as in \cite{liu2019recurrent}. Specifically, standard Average Precision (AP) and Average Recall (AR) are taken as the quantitative metrics. mAP is computed by averaging the results from AP.50 to AP.95 (with a stride 0.05). 

As for the evaluation of counting performance, mean absolute error (MSE) and root mean squared error (MSE) are applied, which are computed as follows,
\begin{align}
	&\mathrm{MAE} = \frac{1}{M}\sum_{i=1}^M{\left|\hat{C}_i-C_i \right| },\\
	&\mathrm{MSE}=  \sqrt{\frac{1}{M}\sum_{i=1}^M{\left(\hat{C}_i-C_i \right)^2 }},
\end{align} 
% \begin{equation}
% \mathrm{MAE} = \frac{1}{N}\sum_{i=1}^N{\left|\hat{C}_i-C_i \right| }, \mathrm{MSE}=  \sqrt{\frac{1}{N}\sum_{i=1}^N{\left(\hat{C}_i-C_i \right)^2 }},
% \end{equation} 
where $\hat{C}_i,C_i$ are the estimated head count and the ground truth one for the $i$-th image, and $M$ is the number of images.

\subsection{Comparison with State-of-the-art Methods}
In this section, we compare our method with existing crowd localization methods on VSCrowd, FDST \cite{fang2019locality}, Venice \cite{liu2019context}, ShanghaiTech \cite{zhang2016single}, and UCF-QNRF \cite{idrees2018composition}, respectively.

\noindent\textbf{1) Localization on VSCrowd Dataset}

In this subsection, we compare GNANet with previous typical crowd localization methods (the localization branch in the main net of RAZNet \cite{liu2019recurrent}, the localization branch of STANet \cite{wen2019drone} and P2PNet \cite{song2021rethinking}) and classic object detection methods (Faster-RCNN \cite{ren2015faster} and FCOS \cite{tian2019fcos}) on the proposed VSCrowd dataset. RAZNet is a single-image crowd localization network that predicts locations of heads in a given image. STANet is the only video crowd localization method that estimates the localization results via a video clip. However, it is specifically designed for drone-view crowd videos. \new{P2PNet is a purely point-based framework for crowd localization, which directly predicts point proposals representing head centers.} For Faster-RCNN and FCOS, they are trained using the head-shoulder box annotations provided in VSCrowd, and the box centers are regarded as the head centers during testing. We have also evaluated the localization performance of the state-of-the art crowd counting method, Generalized Loss \cite{wan2021generalized}. The localization results are obtained by finding the local peaks on the predicted density maps. The results are in TABLE \ref{main}. 

\begin{table}[tbp] 
	\caption{Comparison of localization on VSCrowd. $\sigma$ is the positive range when evaluating the detected points.}
	\label{main}
	\begin{center}
		\resizebox{\columnwidth}{!}{
			\begin{tabular}{lccccccc}
				\hline
				Methods & $\sigma$ & AP.50 & AP.75 & mAP   & AR.50 & AR.75 & mAR   \\ \hline
				& 5    &0.142&0.040&0.058&0.320&0.168&0.180 \\
				Faster R \cite{ren2015faster}& 20    &0.613 &0.500&0.472&0.652&0.593&0.568 \\
				& 40    &0.662 &0.641&0.616&0.678&0.667&0.653 \\ \hline
				& 5     &0.050 &0.018&0.021&0.132&0.060&0.069\\
				FCOS \cite{tian2019fcos}& 20    &0.363&0.274&0.259&0.388&0.338& 0.319\\
				& 40    &0.401 &0.386&0.367&0.406&0.400&0.388 \\ \hline
                 & 5 & 0.069 &0.019  &0.028  &0.213  &0.097 &0.111  \\
                GLoss \cite{wan2021generalized}& 20 & 0.573&0.505&0.446&0.649&0.606&0.554 \\
                 & 40 & 0.605&0.589&0.574&0.665&0.657&0.646 \\ \hline
				& 5     & 0.615 & 0.461  & 0.435 & 0.752 & 0.684  & 0.613 \\
				RAZNet \cite{liu2019recurrent}& 20    & 0.752 & 0.724  & 0.714 & 0.870 & 0.855  & 0.844 \\
				& 40    & 0.774 & 0.757  & 0.755 & 0.854 & 0.844  & 0.843 \\ \hline
				% & 5     &  \\
				% S-GNANet& 20     \\
				% & 40    &  \\ \hline
				& 5     & 0.540 & 0.410  & 0.388 & 0.763 & 0.659  & 0.625 \\
				STANet \cite{wen2019drone}& 20    & 0.691 & 0.663  & 0.652 & 0.874 & 0.855  & 0.847 \\
				& 40    & 0.734 & 0.706  & 0.703 & 0.902 & 0.855  & 0.882 \\\hline
				& 5 & 0.442&0.353&0.330&0.538&0.478&0.451\\
                 \new{P2PNet \cite{song2021rethinking}} & 20 & 0.554&0.525&0.522&0.674&0.657&0.653 \\
                 & 40 &0.586&0.568&0.563&0.694&0.682&0.680 \\\hline 
				& 5     & 0.660 & 0.511  & \textbf{0.480} & 0.778 & 0.677  & \textbf{0.641}\\
				GNANet& 20   &0.775  & 0.757  & \textbf{0.747} &0.885  & 0.873  &\textbf{0.867} \\
				& 40    &0.805  & 0.788  & \textbf{0.782} & 0.902 & 0.891  & \textbf{0.890}\\ \hline 
		\end{tabular}}
		
	\end{center}
\end{table}

\begin{figure}[tbp]
	\centering
	\includegraphics[width=\columnwidth]{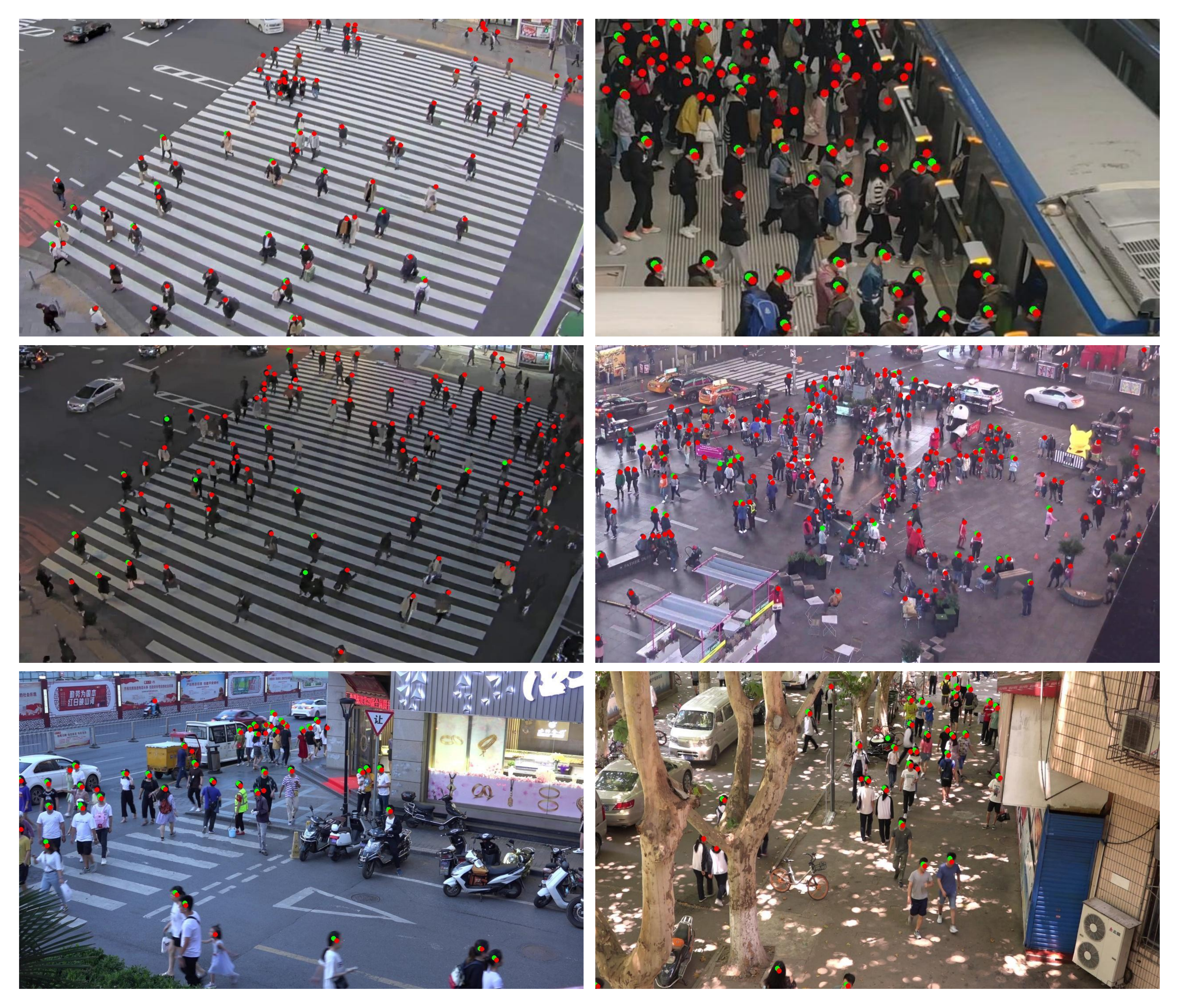}
	\caption{\new{Visualization comparison between GNANet (\textcolor[RGB]{255,0,0}{red} points) and P2PNet (\textcolor[RGB]{0,255,0}{green} points).}}
	\label{comparep2p}
\end{figure}

As shown in TABLE \ref{main}, the object detection methods perform poorly for head center localization when $\sigma$ is small, which means the detection-based methods are not able to locate head centers accurately. When $\sigma$ gets bigger, the two-stage detector Faster R-CNN achieves considerable precision but recall is still poor. The results indicate that the object detection methods can detect the relatively large human heads, while the tiny heads in crowd scenes can not be accurately located. In conclusion, traditional detection-based methods are not capable of handling crowd localization task. As for the counting model trained by Generalized Loss, it shows inferior localization performance, especially under small $\sigma$. The results indicate that counting methods are less capable of locating small heads in crowded or distant regions.

Additionally, the single-image localization method, RAZNet, achieves considerable performance, which sets a strong baseline for the proposed dataset. However, STANet gets poor performance on VSCrowd in terms of both AP and AR. We believe the main reason accounts for these results is that STANet is specifically designed for drone-view crowd scenes whose characteristics are far different from those of surveillance. \new{Besides, the state-of-the-art method, P2PNet, achieves even lower results on our dataset. We assume the reason is that there exists a large distribution gap between the training set and the testing set, and P2PNet cannot handle it since the hyper-parameter used in training is highly related to the property of the dataset.} As for the proposed method, GNANet achieves the best results in both AP and AR under various $\sigma$ values, which means our method can not only find where the human heads are, but also pinpoint the exact locations of head centers. 

Fig. \ref{outputs} illustrates some predictions of RAZNet and GNANet on VSCrowd. As shown in the figure, our method can locate the head centers accurately and handle head size variation well. \new{Apart from RAZNet, we also conduct the visualization comparison the state-of-the-art method, P2PNet, on VSCrowd. The localization predictions are shown in Figure \ref{comparep2p}, where the red/green points are predictions of GNANet/P2PNet. The two images in the first row show the performance of locating small heads in far regions. We can see that P2PNet misses a large proportion of people, while our method detects most of the heads with a high precision. The second row shows the predictions in crowed night scenes. In this situation, P2PNet hardly detects the heads, while our method still performs well. The last row shows the robustness to occlusion and blur. We can see that for some occluded or blurred people, our method can still locate them. In summary, the comparison shows the superiority of our model to P2PNet in dealing with small heads and challenging scenes.}

% \subsubsection{\bf{Localization on FDST Dateset}}
\noindent\textbf{2) Localization on FDST Dataset}

In this subsection, we compare the proposed GNANet with RAZNet, STANet, Generalized Loss, and P2PNet on FDST dataset. FDST is a surveillance dataset for crowd analysis, which contains 15,000 frames and 394,000 annotated heads captured from 13 different scenes. It is similar to VSCrowd but has fewer samples. The results are shown in TABLE \ref{FDST}.

\begin{table}[tbp]
		\caption{Comparison results of localization on FDST. GNANet$^*$ denotes the model pre-trained on VSCrowd.}
	\label{FDST} 
	\begin{center}
		\resizebox{\columnwidth}{!}{
			\begin{tabular}{lccccccc}
				\hline
				Methods & $\sigma$ & AP.50 & AP.75 & mAP   & AR.50 & AR.75 & mAR   \\ \hline
				& 5     & 0.740 & 0.647 & 0.606 & 0.817 & 0.762& 0.606\\
				RAZNet \cite{liu2019recurrent}& 20    &0.797  &0.794  & 0.789 & 0.848 &0.844 & 0.842\\
				& 40    &0.813  &0.808  &  0.805& 0.856 &0.851 &0.851 \\ \hline
				& 5     & 0.738 & 0.636 &0.597  & 0.814 &0.758 &0.723 \\
				STANet \cite{wen2019drone}& 20    & 0.794 & 0.792 &0.786  & 0.846 &0.843 &0.841 \\
				& 40    & 0.808 & 0.796 & 0.799 & 0.853 & 0.848& 0.848\\\hline
                 & 5 & 0.104 & 0.030 & 0.042 & 0.274 & 0.128 & 0.145 \\
                 GLoss \cite{wan2021generalized}& 20 & 0.661 & 0.623 & 0.546 & 0.727 & 0.750 & 0.647 \\
                 & 40 & 0.680 & 0.674 & 0.661 & 0.738 & 0.732 & 0.726\\ \hline 
                 & 5 & 0.630&0.540&0.507&0.826&0.765&0.733\\
                 \new{P2PNet \cite{song2021rethinking}} & 20 & 0.695&0.690&0.683&0.859&0.867&0.861 \\
                 & 40 &0.719&0.705&0.702&0.880&0.871&\textbf{0.870} \\ \hline 
				& 5     & 0.775 &  0.685&  \textbf{0.641}&  0.835& 0.782&\textbf{0.750}\\
				GNANet& 20   & 0.816 & 0.814 & \textbf{0.811} & 0.869 &0.866 &\textbf{0.864} \\
				& 40    & 0.828 & 0.827 & \textbf{0.822} & 0.872 &0.870 &\textbf{0.870}\\ \hline 
				& 5     & 0.790 & 0.698 & \textbf{0.660} & 0.858 &0.806 &\textbf{0.776}\\
				GNANet$^*$& 20   & 0.838 & 0.834 & \textbf{0.829} & 0.896 &0.893 &\textbf{0.891} \\
				& 40    & 0.849 & 0.846 &\textbf{0.842}  & 0.905 &0.909 &\textbf{0.900} \\ \hline 
		\end{tabular}}
	\end{center}
\end{table}

\begin{table}[tbp] 
	\caption{\new{Comparison results of localization on Venice.}}
	\label{venice}
	\begin{center}
		\resizebox{\columnwidth}{!}{
			\begin{tabular}{lccccccc}
				\hline
Methods & $\sigma$ & AP.50 & AP.75 & mAP & AR.50 & AR.75 & mAR \\ \hline
 & 5 & 0.263 & 0.108 & 0.133 & 0.557 & 0.356 & 0.367 \\
RAZNet \cite{liu2019recurrent}& 20 & 0.540 & 0.492 & 0.476 & 0.793 & 0.759 & 0.743 \\
 & 40 & 0.628 & 0.567 & 0.569 & 0.859 & 0.814 & 0.810 \\ \hline
 & 5 & 0.261 & 0.188 & 0.182 & 0.515 & 0.432 & 0.416 \\
STANet \cite{wen2019drone}& 20 & 0.435 & 0.380 & 0.376 & 0.670 & 0.620 & 0.618 \\
 & 40 & 0.554 & 0.481 & 0.476 & 0.754 & 0.701 & 0.689 \\ \hline
 & 5 & 0.354&0.182&\textbf{0.199}&0.670&0.484&\textbf{0.480} \\
GNANet & 20 & 0.536 & 0.496 & \textbf{0.485} & 0.788 & 0.757 & \textbf{0.746} \\
 & 40 & 0.634 & 0.571 & \textbf{0.572} & 0.859 & 0.810 & \textbf{0.812} \\ \hline
 & 5 & 0.597 & 0.451 & \textbf{0.429} & 0.847 & 0.737 & \textbf{0.701} \\
GNANet$^*$ & 20 & 0.690 & 0.665 & \textbf{0.666} & 0.914 & 0.899 & \textbf{0.897} \\
 & 40 & 0.731 & 0.710 & \textbf{0.703} & 0.939 & 0.923 & \textbf{0.922}\\ \hline
		\end{tabular}}
	\end{center}
\end{table}

As shown in TABLE \ref{FDST}, GNANet outperforms other methods (including the localization methods, RAZNet, STANet and P2PNet, and the state-of-the-art counting method, Generalized Loss) to a large extent under various $\sigma$. Note that the performance of all methods on FDST are far higher than those on the proposed VSCrowd, which indicates VSCrowd is much more challenging than FDST. With more scenes and more people, VSCrowd is comprehensive enough to evaluate the performance of crowd analysis methods. 

In addition, to reveal the significance of a large-scale surveillance dataset, we firstly pre-train GNANet on VSCrowd, and then fine-tune it on FDST. The fine-tuned model is denoted as GNANet$^*$. As shown in TABLE \ref{FDST}, the fine-tuned model GNANet$^*$ outperforms GNANet in both AP and AR by a large margin under different values of $\sigma$. The improved performance is expected since a large number of training samples 
endue the model with strong generalizability.

\noindent\textbf{3) Localization on Venice Dataset}

We have also evaluated our method on Venice that is a video dataset for crowd analysis. The results are shown in TABLE \ref{venice}. As we can see from the results, our method outperforms RAZNet and STANet significantly. It is worth noting that Venice is a small dataset consisting limited frames. The results indicate that our method is less prone to overfitting when the training samples are scarce. Furthermore, we pretrain our model on VSCrowd and then fine-tune it on Venice. The fine-tuned model shows great improvement in both precision and recall, which demonstrates the significance of the large-scale dataset of VSCrowd.

\begin{table}[tbp] 
	\caption{\new{Comparisons of localization on DroneCrowd.}}
	\label{DroneCrowd}
	\begin{center}
		\resizebox{\columnwidth}{!}{
			\begin{tabular}{lccccccc}
\hline
Methods & $\sigma$ & AP.50 & AP.75 & mAP & AR.50 & AR.75 & mAR \\ \hline
\multirow{2}{*}{RAZNet \cite{liu2019recurrent}} & 5 & 0.254&0.131&0.144&0.555&0.411&0.404 \\
 & 20 & 0.420&0.392&0.373&0.728&0.694&0.685 \\ \hline
\multirow{2}{*}{P2PNet \cite{song2021rethinking}} & 5 & 0.290&0.174&\textbf{0.181}&0.492&0.378&0.366 \\
 & 20 & 0.442&0.412&0.403&0.604&0.581&0.576 \\ \hline
 \multirow{2}{*}{STANet \cite{wen2019drone}} & 5 & 0.290&0.173&0.179&0.592&0.459&\textbf{0.447} \\
 & 20 & 0.473&0.431&\textbf{0.423}&0.757&0.724&0.715 \\ \hline
\multirow{2}{*}{GNANet} & 5 & 0.250&0.136&0.145&0.590&0.437&0.429 \\
 & 20 & 0.426&0.391&0.380&0.774&0.740&\textbf{0.730} \\ \hline
\end{tabular}}
	\end{center}
\end{table}

\noindent\new{\textbf{4) Localization on DroneCrowd Dataset}\\
We also evaluate our method on DroneCrowd which is a drone-view dataset of crowd videos. Different from FDST, Venice, and our VSCrowd, DroneCrowd are collected in a bird's-eye view, and only the top of the heads can be seen. Without the perspective effect, the heads in the dataset are only several pixels. Therefore, we only report the performance under $\sigma=5$ and $\sigma=20$ in TABLE \ref{DroneCrowd}. The compared methods include RAZNet (a strong baseline method), STANet (specifically designed for DroneCrowd), and P2PNet (the state-of-the-art localization method). Among the compared methods, P2PNet and STANet achieve the best performance in mAP under $\sigma=5$ and $\sigma=20$, respectively. Additionally, STANet and our method achieve the best performance in mAR under $\sigma=5$ and $\sigma=20$, respectively. Even though our model is not specifically designed for drone-view crowd scenes, it still shows considerable performance on this dataset.}

% \subsection{Localization on ShanghaiTech Dateset}
\noindent\textbf{5) Localization on ShanghaiTech Dataset}

\begin{table}[tbp] 
	\caption{Comparison results on ShanghaiTech\_A. The results of RAZNet are reported as in the literature.}
	\label{sha}
	\begin{center}
		\resizebox{\columnwidth}{!}{
			\begin{tabular}{lccccccc}
				\hline
				Methods & $\sigma$ & AP.50 & AP.75 & mAP   & AR.50 & AR.75 & mAR   \\ \hline
				& 5     & 0.308 &0.120   &0.147  &0.529  & 0.326  & 0.331 \\
				RAZNet \cite{liu2019recurrent}& 20    & 0.670 & 0.597  & 0.576 &0.791 & 0.746  &0.729  \\
				& 40    & 0.739 & 0.702  &0.687 &0.836  & 0.811  & 0.803 \\ \hline
				& 5     & 0.380 & 0.191  & \textbf{0.183} & 0.562 & 0.393  &\textbf{0.384} \\
				S-GNANet& 20    & 0.670 & 0.615  & \textbf{0.597} & 0.799 & 0.741  & \textbf{0.740} \\
				& 40    & 0.748 &  0.711 &\textbf{0.699} & 0.842 & 0.819  & \textbf{0.814}\\ \hline
				& 5     &0.391& 0.194 &\textbf{0.210} &0.591 &0.418  &\textbf{0.407}\\
				S-GNANet$^*$& 20    &0.672& 0.619& \textbf{0.611}& 0.843& 0.799 & \textbf{0.786}\\
				& 40    &0.753&  0.719& \textbf{0.710}& 0.859&  0.839&\textbf{0.831}\\ \hline
		\end{tabular}}
		
	\end{center}
\end{table}

\begin{table}[tbp] 
	\caption{Comparison results on ShanghaiTech\_B. The results of RAZNet are reported as in the literature.}
	\label{shb}
	\begin{center}
		\resizebox{\columnwidth}{!}{
			\begin{tabular}{lccccccc}
				\hline
				Methods & $\sigma$ & AP.50 & AP.75 & mAP   & AR.50 & AR.75 & mAR   \\ \hline
				& 5     & 0.356 & 0.156  & 0.181 &0.575  & 0.375  & 0.374 \\
				RAZNet \cite{liu2019recurrent}& 20    & 0.673 & 0.622  & 0.598 &0.793 & 0.761  & 0.746 \\
				& 40    & 0.742 & 0.694  &0.692 & 0.833 & 0.804  & 0.809 \\ \hline
				& 5     &0.421  & 0.211  &\textbf{0.227}  & 0.601 &0.423   & \textbf{0.412}\\
				S-GNANet& 20    &0.688  & 0.625  &  \textbf{0.610}& 0.794 & 0.765 & \textbf{0.768} \\
				& 40    & 0.751 & 0.713  & \textbf{0.701} & 0.848 &  0.826 & \textbf{0.820}\\ \hline
				& 5     & 0.499 & 0.285  &\textbf{0.292}  & 0.675 & 0.507  & \textbf{0.487}\\
				S-GNANet$^*$& 20    &0.718& 0.684 & \textbf{0.668}& 0.821&0.801  & \textbf{0.791} \\
				& 40    & 0.754 & 0.734 &\textbf{0.726} & 0.844&0.831  & \textbf{0.838} \\ \hline
		\end{tabular}}
		
	\end{center}
\end{table}

% \begin{table*}[]
% 	\caption{The counting performance on VSCrowd, FDST and ShanghaiTech (Part A and Part B). Note that the counting results of RAZNet and our methods are derived from the localization results.}
% 	\label{cnt}
% 	\center
% 	\begin{tabular}{lcccccccc}
% 		\hline
% 		\multirow{2}{*}{Methods} & \multicolumn{2}{c}{VSCrowd} & \multicolumn{2}{c}{FDST} & \multicolumn{2}{c}{ShanghaiTech\_A} & \multicolumn{2}{c}{ShanghaiTech\_B} \\
% 		& MAE & MSE & MAE & MSE & MAE  & MSE   & MAE  & MSE  \\ \hline
% 		MCNN \cite{zhang2016single} & 27.1   & 46.9   &  3.8   &   4.9  & 110.2 & 173.2 & 26.4 &41.3 \\
% 		RAZNet \cite{liu2019recurrent}&  12.7   &  19.5   &  4.7 & 5.5  & 75.2 & 133.0 &13.5  & 25.4\\
% 		CSRNet \cite{li2018csrnet} & 13.8    &  21.1   &  ---   &  ---   & 68.2 & 115.0 & 10.6 & 16.0 \\
% 		Bayesian \cite{ma2019bayesian}&  8.7  &  11.8  &--- &--- &64.5& 104.0& 7.9 &13.3 \\\hline
% 		S-GNANet & 9.9 & 12.3  &  3.2  & 4.8  &   70.4   &    124.3   &   11.2   &    22.6\\ \hline
% 		ConvLSTM \cite{xiong2017spatiotemporal}&   28.0  &  46.1   &   4.5  & 5.8    &--- & --- &  ---& --- \\
% 		LST \cite{fang2019locality}&  22.9   &   41.1  &  3.4   &  4.5   & --- & --- & --- &---\\
% 		STANet \cite{wen2019drone}&   12.0  &   18.6  &  3.6   &  5.1   & \textbf{63.7} & \textbf{101.5} & \textbf{7.4} &\textbf{11.0} \\
% 		EPF \cite{liu2020estimating}&   10.4  & 14.6   & 2.2    &   \textbf{2.6}& ---&  ---&  ---&---\\\hline
% 		GNANet & \textbf{8.2} &  \textbf{10.2}   &  \textbf{2.1}  &  2.9   &   ---   &    ---  &   ---   &    ---  \\ \hline
% 	\end{tabular}
% \end{table*}

Apart from performing crowd localization via a video, GNANet can be adapted to a single-image crowd localization model. Specifically, the scene modeling module and temporal GNA in GNANet are discarded, and we directly use the image feature as the scene context and compute self-GNA on it. The attention output and the original feature are combined and sent into the prediction module. Other processes and setups are the same as those of video localization. The single-image-input version of GNANet is denoted as S-GNANet. 
We compare S-GNANet with a typical single-image crowd localization method, RAZNet \cite{liu2019recurrent}, on ShanghaiTech dataset (Part A and Part B) \cite{zhang2016single}. The results are shown in TABLE \ref{sha} (Part A) and TABLE \ref{shb} (Part B).

As shown in TABLE \ref{sha} and TABLE \ref{shb}, S-GNANet achieves better performance than RAZNet under different values of $\sigma$. We also obvious that with the decrease in $\sigma$, the improvement of S-GNANet becomes greater. Since $\sigma$ controls the positive range of detections, a small $\sigma$ means a more strict threshold that a detected head is regarded as a true positive, and a large $\sigma$ makes the improvement less obvious. Therefore, the results indicate S-GNANet can predict the head centers more precisely that RAZNet.

% under $\sigma=5,20$ and comparable performance under $\sigma=40$ on ShanghaiTech\_A. Since $\sigma$ controls the positive range of detections, a small $\sigma$ means a more strict threshold that a detected head is regarded as a true positive. The results indicate S-GNANet is capable of predicting the head centers more precisely that RAZNet. In terms ShanghaiTech\_B, S-GNANet outperforms RAZNet under all $\sigma$ as shown in Table \ref{shb}. 

In addition, to reveal the significance of a large-scale surveillance dataset, we firstly pre-train S-GNANet on the proposed VSCrowd, and then fine-tune it on ShanghaiTech\_A and ShanghaiTech\_B respectively. The fine-tuned model is denoted as S-GNANet$^*$. As shown in TABLE \ref{sha} and TABLE \ref{shb}, S-GNANet$^*$ achieves much higher recall than S-GNANet. Compared to ShanghaiTech\_A, the distribution of ShanghaiTech\_B is more similar to the proposed VSCrowd, so the improvement on ShanghaiTech\_B is more significant than that on ShanghaiTech\_A. We conclude that by being pre-trained on VSCrowd, the recognition of crowd in difficult scenes would be promoted to a large extent.

%\begin{table*}[]
%	\caption{The counting performance on VSCrowd, FDST and ShanghaiTech (Part A and Part B). Note that the counting results of RAZNet and our methods are derived from the localization results.}
%	\label{cnt}
%	\begin{tabular}{lcccccccccc}
%		\hline
%		\multirow{2}{*}{Methods} & \multicolumn{2}{c}{VSCrowd} & \multicolumn{2}{c}{FDST} & \multicolumn{2}{c}{ShanghaiTech\_A} & \multicolumn{2}{c}{ShanghaiTech\_B}& \multicolumn{2}{c}{QNRF} \\
%		& MAE & MSE & MAE & MSE & MAE  & MSE   & MAE  & MSE & MAE & MSE \\ \hline
%		MCNN \cite{zhang2016single} & 27.1   & 46.9   &  3.8   &   4.9  & 110.2 & 173.2 & 26.4 &41.3 &277.0&426.0\\
%		CSRNet \cite{li2018csrnet} & 9.8    &  18.1   &  ---   &  ---   & 68.2 & 115.0 & 10.6 & 16.0 &---&---\\
%		RAZNet \cite{liu2019recurrent}&  8.7   &  16.5   &  4.7 & 5.5  & 75.2 & 133.0 &13.5  & 25.4&135.0&246.0\\\hline
%		ConvLSTM \cite{xiong2017spatiotemporal}&   28.0  &  46.1   &   4.5  & 5.8    &--- & --- &  ---&  ---&---&---\\
%		LST \cite{fang2019locality}&  22.9   &   41.1  &  3.4   &  4.5   & --- & --- & --- & ---&---&---\\
%		STANet \cite{wen2019drone}&   12.0  &   18.1  &  3.6   &  5.1   & \textbf{63.7} & \textbf{101.5} & \textbf{7.4} &\textbf{11.0}&107.6&174.8 \\
%		EPF \cite{liu2020estimating}&   8.1  & 14.6   & 2.2    &   \textbf{2.6}& ---&  ---&  ---&---&---&---\\\hline
%		S-GNANet & 7.9 & 12.1  &  3.2  & 4.8  &   70.4   &    124.3   &   11.2   &    22.6 & &\\ 
%		GNANet & \textbf{6.4} &  \textbf{10.2}   &  \textbf{2.1}  &  2.9   &   ---   &    ---  &   ---   &    ---  & ---& ---\\ \hline
%	\end{tabular}
%\end{table*}

\begin{table}[tbp] 
	\caption{Comparisons of localization on UCF-QNRF.}
	\label{QNRF}
	\begin{center}
		\resizebox{\columnwidth}{!}{
			\begin{tabular}{lccccccc}
\hline
Methods & $\sigma$ & AP.50 & AP.75 & mAP & AR.50 & AR.75 & mAR \\ \hline
 & 5 & 0.135 & 0.047 & 0.062 & 0.286 & 0.158 & 0.172 \\
RAZNet \cite{liu2019recurrent}& 20 & 0.413 & 0.355 & 0.339 & 0.516 & 0.478 & 0.462 \\
 & 40 & 0.477 & 0.443 & 0.431 & 0.555 & 0.533 & 0.526 \\ \hline
  & 5 &0.024&0.005&0.009&0.114&0.049&0.058 \\
\new{GLoss \cite{wan2021generalized}}& 20& 0.411&0.278&0.264&0.520&0.421&0.395\\
 & 40 & 0.520&0.459&0.436&0.581&0.549&0.529\\ \hline
 & 5 & 0.126&0.039&0.055&0.314&0.171&0.185\\
 \new{P2PNet \cite{song2021rethinking}} & 20 & 0.430&0.359&0.344&0.590&0.536&\textbf{0.519} \\
 & 40 & 0.530&0.471&0.462&0.652&0.616&\textbf{0.608}\\ \hline
 & 5 & 0.176&0.061& \textbf{0.082}&0.338&0.195&\textbf{0.207}\\
S-GNANet & 20 & 0.473&0.414&\textbf{0.397}&0.564&0.528&0.513\\
 & 40 & 0.528&0.503&\textbf{0.490}&0.600&0.580&0.574\\ \hline
  & 5 & 0.268 & 0.122 & \textbf{0.142} & 0.496 & 0.327 & \textbf{0.337} \\
S-GNANet$^*$ & 20 & 0.567 & 0.508 & \textbf{0.488} & 0.675 & 0.730 & \textbf{0.690} \\
 & 40 & 0.617 & 0.586 & \textbf{0.576} & 0.763 & 0.745 & \textbf{0.738} \\ \hline
\end{tabular}}
	\end{center}
\end{table}

\begin{table}[tbp] 
	\caption{\new{Comparisons of localization on UCF-QNRF using Precision/Recall/AUC proposed in \cite{idrees2018composition}.}}
	\label{QNRF2}
	\begin{center}
			\begin{tabular}{lccc}
\hline
Methods & Precision &Recall &AUC \\ \hline
MCNN \cite{zhang2016single} & 0.599 &0.635& 0.591  \\
DM-Count \cite{wang2020distribution}& 0.731& 0.638& 0.692  \\
CLoss \cite{idrees2018composition}& 0.758 &0.598 &0.714\\
BLoss \cite{ma2019bayesian}& 0.767 &0.654& 0.720\\
GLoss \cite{wan2021generalized}& 0.787&0.539&0.639\\
P2PNet \cite{song2021rethinking}& 0.812&0.587&0.681\\
\hline
S-GNANet & 0.816&0.674&0.739\\\hline

\end{tabular}
	\end{center}
\end{table}

\begin{table*}[t]
	\caption{The single-image counting performance on VSCrowd, FDST and ShanghaiTech (Part A and Part B). Note that the counting results of RAZNet and our method are derived from the localization results.}
	\label{cnt}
	\center
	\begin{tabular}{lcccccccc}
		\hline
		\multirow{2}{*}{Methods} & \multicolumn{2}{c}{VSCrowd} & \multicolumn{2}{c}{FDST} & \multicolumn{2}{c}{ShanghaiTech\_A} & \multicolumn{2}{c}{ShanghaiTech\_B} \\
		& MAE & MSE & MAE & MSE & MAE  & MSE   & MAE  & MSE  \\ \hline
		MCNN \cite{zhang2016single} & 27.1   & 46.9   &  3.8   &   4.9  & 110.2 & 173.2 & 26.4 &41.3 \\
		RAZNet \cite{liu2019recurrent}&  12.7   &  19.5   &  4.7 & 5.5  & 75.2 & 133.0 &13.5  & 25.4\\
		CSRNet \cite{li2018csrnet} & 13.8    &  21.1   &  ---   &  ---   & 68.2 & 115.0 & 10.6 & 16.0 \\
		Bayesian \cite{ma2019bayesian}&  \textbf{8.7}  &  \textbf{11.8}  &--- &--- &\textbf{64.5}& \textbf{104.0}& \textbf{7.9} &\textbf{13.3} \\\hline
		S-GNANet & 9.9 & 12.3  &  \textbf{3.2}  & \textbf{4.8}  &   70.4   &    124.3   &   11.2   &    22.6 \\ \hline
	\end{tabular}
\end{table*}

\begin{table}[t] 
	\caption{\new{Comparison results of localization on NWPU.}}
	\label{NWPU}
	\begin{center}
		\resizebox{\columnwidth}{!}{
			\begin{tabular}{lccccccc}
\hline
Methods & $\sigma$ & AP.50 & AP.75 & mAP & AR.50 & AR.75 & mAR \\ \hline
 & 5 & 0.158&0.060&0.074&0.300&0.178&0.188 \\
RAZNet \cite{liu2019recurrent}& 20 & 0.385&0.333&0.323&0.480&0.458&0.439\\
 & 40 &  0.436&0.405&0.398&0.515&0.496&0.491\\\hline
  & 5 & 0.009&0.003&0.004& 0.054&0.023&0.028\\
GLoss \cite{wan2021generalized}& 20 &0.263&0.151&0.153&0.327&0.243&0.232 \\
 & 40 & 0.327&0.296&0.273&0.365&0.348&0.329\\\hline
 & 5 & 0.174&0.079&0.092&0.199&0.296&0.199\\
 P2PNet \cite{song2021rethinking} & 20 &0.372&0.335&0.327&0.562&0.533&\textbf{0.526} \\
 & 40 & 0.412&0.388&0.381&0.592&0.575&\textbf{0.570}\\ \hline
 &5 &0.205&0.083&\textbf{0.103} &0.332&0.205&\textbf{0.213}\\
S-GNANet & 20 & 0.418&0.375&\textbf{0.367}&0.482&0.459&0.450 \\
 & 40 & 0.457&0.427&\textbf{0.423}&0.512&0.496&0.492 \\ \hline
\end{tabular}}
	\end{center}
\end{table}

\noindent\textbf{6) Localization on UCF-QNRF Dataset}

\new{We have also evaluated our method on UCF-QNRF that is a larger dataset compared with ShanghaiTech A and B. The results are shown in TABLE \ref{QNRF}. Note that the results of Generalization loss are obtained using the pretrained model provided by the authors. Specifically, Generalization loss achieves less competitive performance under all $\sigma$ compared to the localization methods, which indicates the deficiency of counting methods in predicting exact crowd locations. Additionally, the performance of RAZNet and S-GNANet for detecting small heads is comparable. However, regarding bigger $\sigma$, S-GNANet shows better results of both precision and recall, which means our method is more powerful in dealing with large heads. As for the comparison to P2PNet, our model outperforms it in precision under all $\sigma$, while P2PNet achieves higher recall under bigger $\sigma$. We find that P2PNet tends to have multiple predictions for one head, which can increase the recall in crowded areas but result in more false positives in the scenes containing less people. Furthermore, we pretrain our model on VSCrowd and then fine-tune it on UCF-QNRF. The localization performance is greatly improved.}

\new{Additionally, we use the metric proposed in \cite{idrees2018composition} to evaluate our method. The results are shown in TABLE \ref{QNRF2}. Note that the results of Generalization Loss are obtained using the pretrained model provided by the authors, and the results of P2PNet are reproduced using the official code. As we can see from the reuslts, Generalization Loss and P2PNet achieve competitive average precision, but their average recall is poor compared to our method. Since UCF-QNRF contains a large proportion of small heads, the performance of Generalization Loss and P2PNet indicate their deficiency in detecting small heads.}

\noindent\new{\textbf{7) Localization on NWPU Dataset}}

\new{We have evaluated our method on NWPU \cite{gao2020nwpu} that is a widely used benchmark for crowd analysis in recent works. Since the labels of the testing set are not released, we report the results of the validation set. The proposed method is compared with the strong baseline method, RAZNet, and the state-of-the art counting method, Generalization Loss, and the localization method P2PNet. The results are shown in TABLE \ref{NWPU}. As the results show, our model outperforms other methods significantly in precision. Meanwhile, the results of recall on this dataset are similar to those on UCF-QNRF.}

\subsection{Apply to Crowd Counting}

\begin{table}[]
	\caption{The video counting performance on VSCrowd and FDST. Note that the counting results our method are derived from the localization results.}
	\label{vcnt}
	\center
	\begin{tabular}{lcccc}
		\hline
		\multirow{2}{*}{Methods} & \multicolumn{2}{c}{VSCrowd} & \multicolumn{2}{c}{FDST}  \\
		& MAE & MSE & MAE & MSE  \\ \hline
		ConvLSTM \cite{xiong2017spatiotemporal}&   28.0  &  46.1   &   4.5  & 5.8    \\
		LST \cite{fang2019locality}&  22.9   &   41.1  &  3.4   &  4.5   \\
		STANet \cite{wen2019drone}&   12.0  &   18.6  &  3.6   &  5.1  \\
		EPF \cite{liu2020estimating}&   10.4  & 14.6   & 2.2    &   \textbf{2.6}\\\hline
		GNANet & \textbf{8.2} &  \textbf{10.2}   &  \textbf{2.1}  &  2.9   \\ \hline
	\end{tabular}
\end{table}

\begin{table*}[]
	\caption{The results of ablation study on GNANet. \textit{SM}, \textit{F-GNA}, \textit{T-GNA} and \textit{MF} are the abbreviations for scene modeling module, objective-frame GNA, temporal GNA and multi-focus mechanism respectively. The model in the first row is the baseline that lacks all components, while the last model is GNANet. Note that we only report the results under $\sigma=20$.}
	\label{abl1}
	\begin{center}
		\begin{tabular}{cccc|cccccc}
			\hline
			\textit{SM}   & \textit{F-GNA}   & \textit{T-GNA}  & \textit{MF}  & AP.50 & AP.75 & mAP & AR.50 & AR.75 & mAR \\ \hline
			\multicolumn{4}{c|}{Baseline}  &       0.751   &  0.732   &  0.721     &  0.868      &  0.889 & 0.854 \\\hline
			\checkmark&        &       &     &   0.755    &   0.737     &  0.726   &  0.899     &  0.885      &  0.858   \\
			\checkmark	&   \checkmark     &       &     &   0.763    &   0.745     &  0.735   &   0.884    &  0.872      &  0.863   \\
			\checkmark&        &    \checkmark   &     &   0.761    &   0.742     & 0.731    & 0.885      &  0.871      &  0.864   \\
			\checkmark&   \checkmark     &   \checkmark    &     &   0.767    &   0.749     &  0.739   &  0.874     &  0.862      &   0.865  \\ 
			\checkmark&   \checkmark     &   \checkmark    & \checkmark   &0.775  & 0.757  & \textbf{0.747} &0.885  & 0.873  & \textbf{0.867}    \\ \hline
		\end{tabular}
	\end{center}
\end{table*}

The crowd localization results of our methods can be directly used for crowd counting. In this case, the counting performance of our methods is also evaluated. We compare our methods (GNANet and S-GNANet) with single-image crowd counting methods (MCNN \cite{zhang2016single}, CRSNet \cite{li2018csrnet}, RAZNet \cite{liu2019recurrent} and Bayesian loss \cite{ma2019bayesian}) and video crowd counting methods (ConvLSTM \cite{xiong2017spatiotemporal}, LST \cite{fang2019locality}, STANet \cite{wen2019drone} and EPF \cite{liu2020estimating}) on VSCrowd and two public datasets (FDST and ShanghaiTech). The comparisons of counting performance are shown in TABLE \ref{cnt} and TABLE \ref{vcnt}.

% As shown in Table \ref{cnt}, the proposed methods are comparable with the state of the arts on both video datasets (VSCrowd and FDST) and image datasets (ShanghaiTech\_A and ShanghaiTech\_B) in terms of crowd counting. Specifically,

As shown in TABLE \ref{cnt}, for single-image counting, S-GNANet outperforms MCNN and the localization branch of RAZNet. As for CRSNet and Bayesian loss, they achieve better performance than S-GNANet on Shanghai dataset, but they cannot predict the exact locations of human heads in crowd. On VSCrowd dataset, S-GNANet outperforms CRSNet and achieves sightly poorer performance than Bayesian loss. As shown in TABLE \ref{vcnt}, for video crowd counting, GNANet outperforms other video crowd counting methods on VSCrowd and FDST. By taking temporal information into consideration, GNANet surpasses Bayesian loss and achieves best results among all the compared methods.

% which is based on density regression, and the localization branch of RAZNet, while achieves compare
% slightly poorer performance than CSRNet. The state-of-the-art crowd counting method, Bayesian loss 
% Bayesian loss is a state-of-the-art crowd counting method, which achieves great performance on VSCrowd and Shang 

\begin{table}[tbp] 
	\caption{The results of models with different attention mechanisms. Multi-focus mechanism in GNANet is discarded.}
	\label{abl2}
	\begin{center}
		\resizebox{\columnwidth}{!}{
			\begin{tabular}{lccccccc}
				\hline
				Models & $\sigma$ & AP.50 & AP.75 & mAP   & AR.50 & AR.75 & mAR   \\ \hline
				& 5     & 0.593 & 0.446  & 0.420 & 0.761 & 0.658  &0.623 \\
				RANet & 20    &0.713  &0.694   &0.682  & 0.865 &0.850   &0.845  \\
				& 40    & 0.744 & 0.727  & 0.722 & 0.885 & 0.872  & 0.871\\ \hline
				& 5     &0.628  & 0.470  & 0.443 & 0.765 & 0.656  & 0.620\\
				LANet& 20    & 0.764 &  0.737 & 0.731 & 0.872 &  0.859 & 0.852 \\
				& 40    & 0.783 &  0.767 & 0.765 & 0.833 & 0.825  &0.824 \\ \hline
 & 5 & 0.633 & 0.482 & 0.451 & 0.769 & 0.660 & 0.625 \\
DRANet & 20 & 0.759 & 0.740 & 0.728 & 0.875 & 0.861 & 0.855 \\
 & 40 & 0.790 & 0.773 & 0.768 & 0.892 & 0.881 & 0.875 \\\hline
				& 5     & 0.627 & 0.481  & 0.449 & 0.756 & 0.654  &0.619 \\
				DFANet& 20    &  0.759   & 0.731 &  0.735  &  0.874  & 0.859 & 0.854  \\
				& 40    & 0.779& 0.761 & 0.767 &  0.876&  0.864&0.863\\ \hline
				& 5     & 0.649 & 0.500  & \textbf{0.468} & 0.777 & 0.674  &\textbf{0.638} \\
				GNANet& 20    & 0.767    &   0.749     &  \textbf{0.739}   &  0.874     &  0.862      &   \textbf{0.865}  \\
				& 40    & 0.797 & 0.780  & \textbf{0.774} & 0.891 & 0.881  & \textbf{0.879}\\ \hline
		\end{tabular}}
	\end{center}
\end{table}

\subsection{Ablation Study}

In this subsection, we conduct ablation studies on GNANet. There are four crucial components in GNANet: scene modeling module (\textit{SM}), objective-frame GNA (\textit{F-GNA}), temporal GNA (\textit{T-GNA}) and multi-focus mechanism (\textit{MF}). To verify the effectiveness of each component, we construct simplifications of GNANet as in TABLE \ref{abl1}. Note that the model without any component is the baseline model where the features of all frames are combined by convolution and directly sent into the prediction module. As for the model with \textit{SM} only, the computed scene context is sent into the prediction module. For models without \textit{MF}, $\gamma$ is set to 5. As shown in the table, the baseline model achieves considerable performance, especially in AR. By adding scene modeling, both precision and recall increase incrementally. As for \textit{F-GNA} and \textit{T-GNA}, they both contribute a lot to the performance. Therefore, the context cross-attention module is crucial to GNANet. By comparing the last two rows, we conclude that multi-focus mechanism further improves precision without hurting recall. 

To prove the effectiveness of the proposed Gaussian neighborhood attention (GNA), we construct four variants of GNANet: RANet where GNA in GNANet is replaced with random attention \cite{zaheer2020big}, LANet where GNA is replaced with local attention \cite{ramachandran2019stand} (the neighborhood of a query pixel is defined as a patch of size $9\times9$ centered at the query), DRANet where GNA is replaced with disk random attention (the keys are uniformly sampled from a disk of radius $3\gamma$ centered at the query), and DFANet where GNA is replaced with recent deformable attention \cite{zhu2020deformable}. Multi-focus mechanism in GNANet is abandoned for a fair comparison. The results of the ablation study on attention mechanism are demonstrated in TABLE \ref{abl2}. As shown in the table, RANet achieves poor results in terms of both AP and AR, even poorer than the baseline model in TABLE \ref{abl1}. These results are foreseeable because unrelated and redundant global information is aggregated when computing attention output. As for the model equipped with local attention (LANet), it performs considerably compared with the baseline model in TABLE \ref{abl1}, which means local dependencies are crucial to video crowd localization. Besides, although the disk random attention is similar to the proposed GNA, it only achieves competitive performance to the local attention. By learning the sampling offsets of keys around the query, DFANet outperforms LANet in AR to a great extent when $\sigma$ is large, which indicates the significance of neighborhood deformation. Instead of making the local neighborhood deformable, GNA extends the fixed neighborhood to a random one where the probability of a pixel being a neighbor of a query is determined by their geometrical distance. Being capable of capturing long-range dependencies as well as local correlations, GNA-based GNANet achieves the best results in both AP and AR.

\begin{table}[tbp] 
	\caption{The performance and speed of different methods and GNANet with different numbers of sampling times.}
	\label{sample}
	\begin{center}
		\begin{tabular}{ccccc}
			\hline
			Models & $\sigma$ & mAP    & mAR& FPS  \\ \hline
			RAZNet \cite{liu2019recurrent}& 20  & 0.714 & 0.844 &4.23\\
			STANet \cite{wen2019drone}& 20  & 0.652 & 0.847 &5.34\\\hline
			GNANet(1)& 20  & 0.738($\pm$0.004) &  0.864($\pm$0.006) &3.45\\
			GNANet(3)& 20     & 0.739($\pm$0.001) &  0.865($\pm$0.002) & 3.16\\
			GNANet(5)& 20    & 0.740($\pm$0.001) &  0.867($\pm$0.001) &2.03\\\hline
		\end{tabular}
	\end{center}
\end{table}

\subsection{Further Analysis}
\label{fura}

\textbf{1) Randomness of GNA.} To demonstrate the impact of randomness caused by Gaussian sampling in GNA when testing, we conduct experiments on the multi-sample mechanism where GNA is computed multiple times and the average is taken as the final output of the GNA module. We report the means and standard deviations based on 5 times of testing in TABLE \ref{sample}. Note that the multi-focus mechanism is abandoned in this experiment and the following ones to make the impact evident. As shown in TABLE \ref{sample}, GNANet with different numbers of sampling times achieves similar mAP and mAR in average. However, the deviation of performance decreases with the increase in sampling time. In terms of time consumption, the inference speed drops with more sampling times as expected. To balance the performance stability and efficiency, we choose to sample 3 times in GNA and report the average performance in all experiments. Additionally, we have compared the computational cost of our methods with RAZNet and STANet. As the results show, although GNANet is inferior to RAZNet and STANet in terms of computational cost, it outperforms others significantly in both precision and recall, especially compared with STANet. In conclusion, our method achieves a reasonable balance between localization accuracy and computational cost.

\begin{table}[tbp] 
		\caption{The results of GNANet with different numbers of sampled points in GNA on VSCrowd.}
	\label{sm}
	\begin{center}
		\resizebox{\columnwidth}{!}{
			\begin{tabular}{cccccccc}
				\hline
				\#Points & $\sigma$ & AP.50 & AP.75 & mAP   & AR.50 & AR.75 & mAR   \\ \hline
				& 5    &0.649&0.498&0.456&0.779&0.673&0.627 \\
				8& 20   &0.765&0.749&0.733&0.861&0.849&0.844 \\
				& 40   &0.806&0.789&0.769&0.876&0.866&0.865 \\ \hline
				& 5    &0.624&0.463&0.458&0.793&0.678&0.631 \\
				16& 20   &0.763&0.744&0.733&0.886&0.873&0.861 \\
				& 40    &0.794&0.777&0.772&0.903&0.892&0.871\\ \hline
				& 5     & 0.649 & 0.500  & \textbf{0.468} & 0.777 & 0.674  &\textbf{0.638} \\
				32& 20    & 0.767    &   0.749     &  \textbf{0.739}   &  0.874     &  0.862      &   \textbf{0.865}  \\
				& 40    & 0.797 & 0.780  &\textbf{0.774} & 0.891 & 0.881  & \textbf{0.879}\\ \hline
		\end{tabular}}
	
	\end{center}
\end{table}

\textbf{2) Impact of the Number of Sampled Points.} The number of sampled points is crucial to localization, since it controls the amount of information that is aggregated to a query. TABLE \ref{sm} demonstrates the performance of GNANet with different numbers of sampled points. As shown in the table, with the increase in sampling number, the performance of GNANet is improved significantly. We believe two reasons account for the improvement: a) Sampling  more points means more information to be considered when computing the attention output for a specific query; b) The stability of GNANet is increased since more sampled points alleviates the randomness in GNA. However, considering computation complexity, 32 points are sampled from the Gaussian neighborhood, aiming to achieve balance performance and efficiency.

\begin{table}[tbp] 
		\caption{The results of different values of $\gamma$ and their different combinations in GNA on VSCrowd.}
	\label{gamma}
	\begin{center}
		\resizebox{\columnwidth}{!}{
\begin{tabular}{cccccccc}
\hline
$\gamma$ & $\sigma$ & AP.50 & AP.75 & mAP & AR.50 & AR.75 & mAR \\ \hline
\multirow{3}{*}{3} & 5 & 0.631 & 0.475 & 0.470 & 0.783 & 0.672 & 0.640 \\
 & 20 & 0.770 & 0.743 & 0.736 & 0.782 & 0.859 & 0.853 \\
 & 40 & 0.793 & 0.776 & 0.766 & 0.888 & 0.878 & 0.856 \\ \hline
\multirow{3}{*}{5} & 5 & 0.649 & 0.500 & 0.468 & 0.777 & 0.674 & 0.638 \\
 & 20 & 0.767 & 0.749 & 0.739 & 0.874 & 0.862 & 0.865 \\
 & 40 & 0.797 & 0.780 & 0.774 & 0.891 & 0.881 & 0.879 \\ \hline
\multirow{3}{*}{10} & 5 & 0.622 & 0.475 & 0.448 & 0.807 & 0.698 & 0.631 \\
 & 20 & 0.754 & 0.728 & 0.723 & 0.892 & 0.879 & \textbf{0.874} \\
 & 40 & 0.776 & 0.759 & 0.780 & 0.909 & 0.898 & 0.887 \\ \hline
\multirow{3}{*}{15} & 5 & 0.631 & 0.485 & 0.433 & 0.778 & 0.673 & 0.627 \\
 & 20 & 0.757 & 0.739 & 0.732 & 0.857 & 0.846 & 0.841 \\
 & 40 & 0.787 & 0.770 & 0.767 & 0.872 & 0.863 & 0.861 \\ \hline
\multirow{3}{*}{3/5/10} & 5 & 0.660 & 0.511 & \textbf{0.480} & 0.778 & 0.677 & \textbf{0.641} \\
 & 20 & 0.775 & 0.757 & 0.747 & 0.885 & 0.873 & 0.867 \\
 & 40 & 0.805 & 0.788 & \textbf{0.782} & 0.902 & 0.891 & 0.890 \\ \hline
\multirow{3}{*}{3/5/15} & 5 & 0.643 & 0.483 & 0.454 & 0.776 & 0.665 & 0.628 \\
 & 20 & 0.771 & 0.754 & 0.745 & 0.858 & 0.846 & 0.841 \\
 & 40 & 0.801 & 0.785 & 0.783 & 0.873 & 0.864 & 0.863 \\ \hline
\multirow{3}{*}{3/10/15} & 5 & 0.628 & 0.471 & 0.441 & 0.789 & 0.674 & 0.638 \\
 & 20 & 0.765 & 0.747 & 0.739 & 0.879 & 0.866 & 0.860 \\
 & 40 & 0.795 & 0.779 & 0.777 & 0.896 & 0.886 & 0.884 \\ \hline
\multirow{3}{*}{5/10/15} & 5 & 0.615 & 0.462 & 0.433 & 0.797 & 0.680 & 0.639 \\
 & 20 & 0.775 & 0.757 & \textbf{0.749} & 0.877 & 0.865 & 0.859 \\
 & 40 & 0.782 & 0.765 & 0.766 & 0.909 & 0.898 & \textbf{0.896} \\ \hline
\end{tabular}}
	\end{center}
\end{table}

\textbf{3) Impact of Standard Deviation in GNA.} The standard deviation $\gamma$ of Gaussian distribution in GNA plays an important role in modeling human heads in crowd videos, which controls the range of information aggregation. To demonstrate the impact of $\gamma$, we conduct experiments on it and the results are shown in TABLE \ref{gamma}. As shown in the table, GNANet with a large/small $\gamma$ achieves high performance under a large/small $\sigma$, which means different values of $\gamma$ focus on different sizes of heads. However, when $\gamma=15$, the performance drops considerably under various $\sigma$. We believe that $\gamma=15$ is too big for head modeling in VSCrowd, so redundant information is aggregated into a head center when computing the attention output. Additionally, we also compare different combinations of $\gamma$ for GNA in TABLE \ref{gamma}. As we can see from the results, the multi-focus mechanism improves the performance of video crowd localization in terms of both precision and recall. Specifically, the combination of 3/5/10 shows a good balance between mAP and mAR under various $\sigma$. In this case, we choose to use the combination of 3/5/10 for the multi-focus GNA.

%	More comparisons and ablation studies can be found in the supplemental material. 

\section{Conclusion}
\label{con}

In this paper, we explore the crucial video crowd localization task which aims to locate human heads in the objective frame of a video. Since there exist limited large-scale surveillance datasets for video crowd analysis, we create a benchmark called VSCrowd, which is a comprehensive and large-scale dataset for this task. To exploit the spatial-temporal dependencies in a video, we design a Gaussian neighborhood attention (GNA), which is capable of capturing long-range correspondences while maintaining the spatial topology structure in the data. Equipped with the multi-focus mechanism, GNA can model the scale variation of human heads in crowd scenes. Based on multi-focus GNA, we develop an end-to-end network structure called GNANet to accurately locale human heads in videos, where a scene modeling module and a context cross-attention module are leveraged to aggregate spatial-temporal information. Extensive experiments on VSCrowd and two public datasets have proved the superiority of GNANet and the effectiveness of the proposed modules for video crowd localization and counting.

\newcommand{\norm}[1]{\left\lVert#1\right\rVert}
\appendices
\section{Theoretical Analysis on Sparse Attention}\label{ta}
Assuming $\bm{K},\bm{V}\in\mathbb{R}^{n\times d}$ are the key matrix and the value matrix for a specific query $\bm{q}\in\mathbb{R}^{d}$, respectively, the output of the full attention (FA) for the query $\bm{q}$ is computed as
\begin{equation}
	\mathrm{FA}(\bm{q},\bm{K},\bm{V})=\mathrm{SM}\left(\bm{q}\bm{K}^\mathrm{T}\right)\bm{V}\in \mathbb{R}^{d},
\end{equation}
where $\mathrm{SM}(\cdot)$ is the softmax operation, and dot-product is used as the score function. For the sparse attention, we select only a small number of keys whose index set is given as $\mathcal{P}$. Then, the key matrix $\bm{K}_1\in\mathbb{R}^{n\times d}$ and the value matrix $\bm{V}_1\in\mathbb{R}^{n\times d}$ for the sparse attention are constructed as follows,
\begin{equation}
(\bm{K}_1)_{ij}=\left\{\begin{matrix}\bm{K}_{ij},&\mathrm{if}\ i\in \mathcal{P},\\0,&\mathrm{otherwise,}\\\end{matrix}\right.
\end{equation}
\begin{equation}
(\bm{V}_1)_{ij}=\left\{\begin{matrix}\bm{V}_{ij},&\mathrm{if}\ i\in \mathcal{P},\\0,&\mathrm{otherwise}.\\\end{matrix}\right.
\end{equation}
Similarly, the output of the sparse attention (SA) for the query $\bm{q}$ is computed as
\begin{equation}
	\mathrm{SA}(\bm{q},\bm{K}_1,\bm{V}_1)=\mathrm{SM}\left(\bm{q}\bm{K}_1^\mathrm{T}\right)\bm{V}_1\in \mathbb{R}^{d},
\end{equation} 
For clarity, we define the following auxiliary notions,
\begin{align}
   &C=\norm{\mathrm{SM}\left(\bm{q}\bm{K}^\mathrm{T}\right)},\\
   &\bm{K}_2=\bm{K}-\bm{K}_1,\\
   &\bm{V}_2=\bm{V}-\bm{V}_1,
\end{align}
where $C$ is a constant irrelevant to the sparse attention, $\bm{K}_2,\bm{V}_2$ are constructed by the unselected keys and values, respectively.
Then the error of estimating the full attention by the sparse attention is derived as follows,
\begin{align*}
&\norm{\mathrm{FA}(\bm{q},\bm{K},\bm{V})-\mathrm{SA}(\bm{q},\bm{K}_1,\bm{V}_1)}\\
=&\norm{\mathrm{SM}\left(\bm{q}\bm{K}^\mathrm{T}\right)\bm{V}-\mathrm{SM}\left(\bm{q}\bm{K}_1^\mathrm{T}\right)\bm{V}_1}\\
=&\norm{\mathrm{SM}\left(\bm{q}\bm{K}^\mathrm{T}\right)(\bm{V}_1+\bm{V}_2)-\mathrm{SM}\left(\bm{q}\bm{K}_1^\mathrm{T}\right)\bm{V}_1}\\
=&\norm{\mathrm{SM}\left(\bm{q}\bm{K}^\mathrm{T}\right)\bm{V}_1+\mathrm{SM}\left(\bm{q}\bm{K}^\mathrm{T}\right)\bm{V}_2-\mathrm{SM}\left(\bm{q}\bm{K}_1^\mathrm{T}\right)\bm{V}_1}\\
\leq&\norm{\mathrm{SM}\left(\bm{q}\bm{K}^\mathrm{T}\right)\bm{V}_1-\mathrm{SM}\left(\bm{q}\bm{K}_1^\mathrm{T}\right)\bm{V}_1}+\norm{\mathrm{SM}\left(\bm{q}\bm{K}^\mathrm{T}\right)\bm{V}_2}\\
\leq&\norm{\mathrm{SM}\left(\bm{q}\bm{K}^\mathrm{T}\right)-\mathrm{SM}(\bm{q}\bm{K}_1^\mathrm{T})}\norm{\bm{V}_1}+\norm{\mathrm{SM}(\bm{q}\bm{K}^\mathrm{T})}\norm{\bm{V}_2}\\
\leq&\norm{\bm{q}\bm{K}^\mathrm{T}-\bm{q}\bm{K}_1^\mathrm{T}}\norm{\bm{V}_1}+C\norm{\bm{V}_2}\\
=&\norm{\bm{q}\bm{K}_2^\mathrm{T}}\norm{\bm{V}_1}+C\norm{\bm{V}_2}.
\end{align*}

As we can see from the upper bound of the error, it is partially decided by the the dot-product (similarity) between the query $\bm{q}$ and those keys that are NOT participated in the computation of the attention. Statistically, the similarity between two features at different positions is negatively correlated to their geometrical distance. In this case, compared to the random attention, the proposed GNA can estimate the full attention with smaller error expectation since a query attends to only the keys sampled from its neighborhood in GNA.

\bibliographystyle{IEEEtran}
\bibliography{ref}

\begin{IEEEbiographynophoto}{Haopeng Li}
is a Ph.D student in the School of Computing and Information Systems, University of Melbourne. He received his B.S. degree in the School of Science, Northwestern Polytechnical University, and the Master degree in the School of Artificial Intelligence, Optics and Electronics (iOPEN), Northwestern Polytechnical University. His research interests include computer vision, video understanding, and artificial intelligence.
\end{IEEEbiographynophoto}

\begin{IEEEbiographynophoto}{Lingbo Liu}
received the Ph.D degree from the School of Computer Science and Engineering, Sun Yat-sen University, Guangzhou, China, in 2020. From March 2018 to May 2019, he was a research assistant at the University of Sydney, Australia. His current research interests include machine learning and urban computing. He has authorized and co-authorized on more than 15 papers in top-tier academic journals and conferences. He has been serving as a reviewer for numerous academic journals and conferences such as TPAMI, TKDE, TNNLS, TITS, CVPR, ICCV and IJCAJ.
\end{IEEEbiographynophoto}

\begin{IEEEbiographynophoto}{Kunlin Yang}
is an algorithm engineer in SenseTime Group Limited.
\end{IEEEbiographynophoto}
\begin{IEEEbiographynophoto}{Shinan Liu}
is an algorithm engineer in SenseTime Group Limited.
\end{IEEEbiographynophoto}

\begin{IEEEbiographynophoto}{Junyu Gao}
received his master and Ph.D degrees from Center for Optical Imagery Analysis and Learning, Northwestern Polytechnical University, advised by Qi Wang, Yuan Yuan and Xuelong Li. His research interests are in computer vision, deep learning and machine learning.
\end{IEEEbiographynophoto}

\begin{IEEEbiographynophoto}{Bin Zhao}
is a doctor with the School of Artificial Intelligence, Optics and Electronics (iOPEN), Northwestern Polytechnical University, Xi’an 710072, P. R. China. His research interests include physics models and cognitive science to artificial intelligence.
\end{IEEEbiographynophoto}

\begin{IEEEbiographynophoto}{Rui Zhang}
(\url{www.ruizhang.info}) is a visiting professor at Tsinghua University and was a Professor at the University of Melbourne. His research interests include AI and big data, particularly in the areas of recommender systems, knowledge bases, chatbot, and spatial and temporal data analytics. Professor Zhang has won several awards including Future Fellowship by the Australian Research Council in 2012, Chris Wallace Award for Outstanding Research by the Computing Research and Education Association of Australasia in 2015, and Google Faculty Research Award in 2017.
\end{IEEEbiographynophoto}

\begin{IEEEbiographynophoto}{Jun Hou}
is with SenseTime Group Limited.
\end{IEEEbiographynophoto}

\end{document}